\PassOptionsToPackage{numbers}{natbib}  
\documentclass[manuscript,screen]{acmart}

\usepackage{url}
\usepackage{verbatim}
\usepackage{graphicx}
\usepackage{xcolor}
\usepackage{colortbl}

\usepackage{amsmath}
\usepackage{mathtools}
\usepackage{amsthm}

\usepackage{algorithm}
\usepackage{algpseudocode}
\usepackage{booktabs}

\theoremstyle{plain}

\theoremstyle{definition}

\theoremstyle{remark}

\usepackage[textsize=tiny]{todonotes}
\usepackage{ulem}

\usepackage[inline]{enumitem}
\usepackage{multirow}
\graphicspath{{./img/}}
\usepackage{xspace}
\DeclarePairedDelimiterX{\infdivx}[2]{(}{)}{%
	#1\;\delimsize|\delimsize|\;#2%
}
\newcommand{\kld}[2]{\ensuremath{D_{KL}\infdivx{#1}{#2}}\xspace}

\newcommand{\modelname}{MTabGen}
\newcommand{\xcond}{x^\text{cond}}
\newcommand{\xmask}{x^\text{mask}}

\newcount\Comments  % 0 suppresses notes to selves in text
\AtBeginDocument{%
	}

\setcopyright{acmlicensed}
\copyrightyear{2025}
\acmYear{2025}
\acmDOI{XXXXXXX.XXXXXXX}

\acmISBN{978-1-4503-XXXX-X/24/05}

\begin{document}
	
	\title{Diffusion Models for Tabular Data Imputation and Synthetic Data Generation}
	
	\author{Mario Villaizán-Vallelado}
	\email{mario.villaizan@uva.es}
	\orcid{0009-0002-0754-1742}
	\affiliation{%
		\institution{Universidad de Valladolid}
		\city{Valladolid}
		\country{Spain}
	}
	\affiliation{%
		\institution{Telefónica Scientifc Research}
		\city{Madrid}
		\country{Spain}}
	
	\author{Matteo Salvatori}
	\orcid{0000-0003-1499-6024}
	\affiliation{%
		\institution{Telefónica Scientifc Research}
		\city{Madrid}
		\country{Spain}}
	\email{matteo.salvatori@telefonica.com}
	
	\author{Carlos Segura}
	\orcid{0000-0001-5867-281X}
	\affiliation{%
		\institution{Telefónica Scientifc Research}
		\city{Madrid}
		\country{Spain}}
	\email{carlos.seguraperales@telefonica.com}
		
	\author{Ioannis Arapakis}
	\orcid{0000-0003-2528-6597}
	\affiliation{%
		\institution{Telefónica Scientifc Research}
		\city{Madrid}
		\country{Spain}}
	\email{ioannis.arapakis@telefonica.com}
	
	\renewcommand{\shortauthors}{Villaizán-Vallelado et al.}
	

\begin{abstract}

Data imputation and data generation have important applications across many domains where incomplete or missing data can hinder accurate analysis and decision-making.
Diffusion models have emerged as powerful generative models capable of capturing complex data distributions across various data modalities such as image, audio, and time series. Recently, they have been also adapted to generate tabular data. In this paper, we propose a diffusion model for tabular data that introduces three key enhancements: (1) a conditioning attention mechanism, (2) an encoder-decoder transformer as the denoising network, and (3) dynamic masking.
The conditioning attention mechanism is designed to improve the model's ability to capture the relationship between the condition and synthetic data. The transformer layers help model interactions within the condition (encoder) or synthetic data (decoder), while dynamic masking enables our model to efficiently handle both missing data imputation and synthetic data generation tasks within a unified framework.
We conduct a comprehensive evaluation by comparing the performance of diffusion models with transformer conditioning against state-of-the-art techniques such as Variational Autoencoders, Generative Adversarial Networks and Diffusion Models, on benchmark datasets. 
Our evaluation focuses on the assessment of the generated samples with respect to three important criteria, namely: (1) Machine Learning efficiency, (2) statistical similarity, and (3) privacy risk mitigation. For the task of data imputation, 
we consider the efficiency of the generated samples across different levels of missing features.
The results demonstrates average superior machine learning efficiency and statistical accuracy compared to the baselines, while maintaining privacy risks at a comparable level, particularly showing increased performance in datasets with a large number of features.
By conditioning the data generation on a desired target variable, the model can mitigate systemic biases, generate augmented datasets to address data imbalance issues, and improve data quality for subsequent analysis.
This has significant implications for domains such as healthcare and finance, where accurate, unbiased, and privacy-preserving data are critical for informed decision-making and fair model outcomes \footnote{Source code will be made available upon acceptance of the manuscript}.

\end{abstract}

\ccsdesc[300]{Computer systems organization~Neural networks}
\ccsdesc[100]{Computing methodologies~Temporal difference learning}
\ccsdesc[500]{Computing methodologies~Learning latent representations}
\ccsdesc[100]{Computing methodologies~Machine learning approaches}
\ccsdesc[100]{Computing methodologies~Machine learning}
\ccsdesc[100]{Computing methodologies}
\ccsdesc[300]{Computing methodologies~Neural networks}
	
	\keywords{Data imputation, synthetic data generation, Diffusion Model, Generative Model, Transformer}
	
	\received{20 February 2007}
	\received[revised]{12 March 2009}
	\received[accepted]{5 June 2009}
	
	\maketitle

\section{Introduction\label{sec:introduction}}

The exponential increase in data generation across sectors such as healthcare, finance, telecommunications, and energy has significantly enhanced decision-making capabilities powered by Artificial Intelligence (AI) and Machine Learning (ML) technologies. However, the presence of missing or incomplete data poses significant challenges, undermining the reliability of analyses derived from ML algorithms. Moreover, the rise in strict AI regulations and data protection laws has intensified the need for robust data privacy measures, challenging traditional data handling practices.

Centralised, cloud-based ML solutions, while efficient in terms of model performance, have been repeatedly criticised for their inherent data privacy issues. Specifically, the centralised nature of these solutions necessitates the transfer of large volumes of multidimensional and privacy-sensitive user data, which raises significant privacy concerns. Moreover, centralised models contend with the issue of single point failure. 
Even advanced solutions like Federated Learning (FL), which aim to decentralize data processing to enhance privacy, depend on a central server for coordinating training processes and aggregating updates. This centralization leaves systems vulnerable to potential privacy risks from information leakage attacks \cite{Nasr2019Comprehensive,Hitaj2017Deep}, which can infer private data from shared model updates.

In addition, these ML models are particularly prone to utility loss due to missing or corrupted data, especially when dealing with sparse datasets.  The handling of missing data can favour certain statistical interpretations, and subsequent implications for policy and practice \cite{Cox2014Working}. For instance, the mean substitution method, often used to handle missing data, may lead to inconsistent bias, especially in the presence of a great inequality in the number of missing values for different features \cite{Kang2013prevention}. Furthermore, when missing data are not missing at random (MNAR), even multiple imputations do not lead to valid results \cite{Twisk2022Handling}. Such imputation methods that fill in blanks with estimated values may inadvertently lead to the creation and transmission of inaccurate or misleading information.

Current efforts to address these issues, such as the application of differential privacy or the use of specialised hardware (e.g., Trusted Execution Environments \cite{Mo2021PPFL}), often result in a trade-off between privacy and data utility or necessitate additional infrastructure. 

Given these challenges, there is a pressing need for solutions that effectively manage data integrity and privacy. Recent advancements in machine learning, specifically Generative Adversarial Networks (GANs) \citep{goodfellow2014generative} and Diffusion Models \citep{sohl2015deep, ho2020denoising}, have shown promise in generating high-fidelity synthetic data  that preserve the statistical properties of original datasets while mitigating privacy concerns, since they follow the original distribution without directly exposing or replicating sensitive information. These generative methods have found their way into applications like image and audio processing \citep{zhang2022styleswin, rombach2022high, liu2023audioldm} and, more recently, have expanded to address tabular data as well \citep{xu2019modeling, kotelnikov2022tabddpm,kim2023stasy, lee2023codi,zhang2024mixedtype}.

Specifically, for tabular data, synthetic data stands out as a privacy-preserving alternative to real data that may contain personally identifiable information. It enables the generation of datasets that mimic the statistical properties of their original counterparts, while mitigating the risk of individual privacy breaches. 
In addition, this approach to generating new samples can augment existing datasets by, for example, correcting class imbalances, reducing biases, or expanding their size when dealing with sparse or limited data. Furthermore, integrating methods for differential privacy~\citep{dwork2006differential,jalko2021privacy} with generative models for tabular data makes possible to share synthetic datasets across units in large organizations, addressing legal or privacy concerns that often impede technological innovation adoption.

In this study, we consider synthetic data generation as a general case of data imputation. In instances where every column in a sample has missing values, the task of data imputation naturally transitions to synthesizing new data. We introduce \textit{\modelname{}}, a new conditioning in diffusion model for tabular data using an encoder-decoder transformer and a dynamic masking mechanism that makes it possible to tackle both tasks with a single model. 
During the training step of the model, the dynamic masking randomly masks features that we later generate or impute. The unmasked features are used as context or condition for the reconstruction of masked features during the reverse denoising phase of the diffusion process. 
We refer to the features to be denoised during the reverse denoising phase as \textit{masked features} hereinafter.

In our analysis, we perform a rigorous evaluation of the proposed approach using several benchmark datasets, each with a wide range of features. 
We demonstrate that our method shows overall improved performance compared to existing baselines, particularly in handling high-dimensional datasets, thereby highlighting its robustness and adaptability in complex tabular data scenarios. The key contributions of this work are the following:

\begin{itemize}
\itemsep0em
\item We propose a new conditioning mechanism for tabular diffusion models. We model the interaction between condition and masked features (e.g., features to-be-denoised) by using an attention mechanism. Within this mechanism, the embedding of the masked features plays the role of query ($\bf{Q}$), and the embedding of the condition plays the roles of key ($\bf{K}$) and value ($\bf{V}$) (see  \cite{vaswani2017attention} for details). Compared to the standard approach, where condition and masked features are added \citep{zheng2022diffusion, kotelnikov2022tabddpm} or concatenated \citep{lee2023codi}, our method allows to learn more complex relationships showing globally improved performances.
\item We incorporate a full encoder-decoder transformer within the diffusion process as the denoising model. The encoder learns the condition embedding, while the decoder models the representation of the masked features. Using transformer layers enhances the learning of inter-feature interactions: within the condition  for the encoder and the masked features for the decoder. Additionally, the encoder-decoder architecture allows the implementation of the conditioning attention mechanism explained in the previous item. To the best of our knowledge, \citep{zheng2022diffusion} is the only prior work that has considered diffusion models for tabular data, using a transformer denoising component. However, in this paper, the transformer layer is limited to learning the masked feature representation, without modeling the condition nor using conditioning attention mechanism. 
\item We extend the masking mechanism proposed by \citet{zheng2022diffusion} to train a single model capable of multitasking, handling both missing data imputation and synthetic data generation. This is facilitated by the transformer encoder-decoder architecture, which allows for arbitrary modification of the split between condition and masked features during the training phase.
\item We conduct extensive experiments on several public datasets and demonstrate average performance gain over state-of-the-art baselines, for both tasks, missing data imputation and synthetic data generation. We evaluate the synthetic data with respect to three important criteria: (1) ML efficiency, (2) statistical similarity and (3) privacy risk.
\end{itemize}

\section{Related work\label{sec:Related_works}}

\textbf{Tabular Data Modeling and Benchmarks.}
The evaluation of models for tabular data requires diverse benchmarks that reflect real-world data conditions. Several recent studies have established such benchmarks for both data imputation and supervised learning. \citet{jager2021benchmark} present a comprehensive evaluation of data imputation methods, comparing both classical and deep learning-based imputation approaches. Their results show that imputation can significantly improve downstream ML tasks, especially when training data is fully observed. For supervised tasks, \citet{grinsztajn2024why} argue that tree-based models such as XGBoost and Random Forests often outperform deep learning models due to the specific properties of tabular data, including irregular patterns and uninformative features. The authors provide a standard benchmark for evaluating both traditional and novel approaches on a variety of real-world datasets. \citet{borisov2024deep} also highlight the challenges faced by deep learning models when dealing with heterogeneous tabular data and emphasize the need for architectures specifically designed for tabular datasets. In this work, we extend the benchmark proposed by \citep{borisov2024deep} to include a wider variety of datasets, providing a more exhaustive evaluation of generative tasks.

\textbf{Diffusion Models.}
Originally introduced by \citet{sohl2015deep} and \citet{ho2020denoising}, diffusion models utilize a two-step generative approach. Initially, they degrade a data distribution using a forward diffusion process by continuously introducing noise from a known distribution. Subsequently, they employ a reverse process to reconstruct the original data structure. At their core, these models leverage parameterized Markov chains, starting typically from a foundational distribution such as a standard Gaussian, and use deep neural networks to reverse the diffusion. Demonstrated by recent advancements \citep{nichol2021improved, dhariwal2021diffusion}, diffusion models have showcased their capability, potentially surpassing GANs in image generation capabilities. Recent works, such as \textit{StaSy} \citep{kim2023stasy}, \textit{CoDi} \citep{lee2023codi}, \textit{Tabsyn} \citep{zhang2024mixedtype}, \textit{TabDDPM} \citep{kotelnikov2022tabddpm} or \textit{TabCSDI} \citep{zheng2022diffusion}, adapt diffusion models to handle tabular data in both synthetic data generation and data imputation tasks. 

\textbf{Missing Data Imputation.}
Handling missing values in datasets is a non-trivial problem. Traditional approaches may involve excluding rows or columns with missing entries or imputing missing values using the average values of the corresponding feature. However, recent efforts have been focusing on ML techniques \citep{van2011mice,bertsimas2017predictive,jarrett2022hyperimpute} and deep generative models \citep{yoon2018gain,biessmann2019datawig,wang2021pc,ipsen2022deal,kyono2021miracle}, and new models using diffusion processes have been developed for data imputation tasks. Specifically, \textit{TabCSDI}, based on the \textit{CSDI} model originally designed for time-series data, adapts this technology for tabular data imputation. \textit{TabCSDI} employs three common preprocessing techniques: 
\begin{enumerate*}[label={(\arabic*)}]
\item one-hot encoding, 
\item analog bits encoding, and 
\item feature tokenization
\end{enumerate*}. These methods allow it to treat continuous and categorical variables uniformly in a Gaussian diffusion process, regardless of their original types. In contrast,  \textit{\modelname{}} implements a dual diffusion mechanism adapted for both continuous and categorical data, maintaining the unique statistical properties of each feature type throughout the diffusion process. As noted previously, \textit{TabCSDI} uses a transformer layer to learn only the interactions within the masked features and then adds the transformer output to the condition embedding. In our case, \textit{\modelname{}} uses a more adaptable and general approach: the condition embedding is modeled by using a transformer encoder and then fed into a conditioning attention mechanism.

\textbf{Generative models.} The application of this family of models to tabular data has been gaining increased attention within the ML community. In particular, tabular VAEs \citep{xu2019modeling} and GANs 
\citep{engelmann2021conditional, torfi2022differentially, zhao2021ctabgan, kim2021oct, zhang2021ganblr, nock2022generative, wen2022causal} 
have shown promising results. Recently, \textit{StaSy} \citep{kim2023stasy}, \textit{CoDi} \citep{lee2023codi}, \textit{Tabsyn} \citep{zhang2024mixedtype}, \textit{TabDDPM} \citep{kotelnikov2022tabddpm} have been proposed as powerful alternatives to tabular data generation, leveraging the strengths of Diffusion Models. 

Specifically, \textit{STaSy}, applies a score-based generative approach \citep{song2021scorebased}, integrating self-paced learning and fine-tuning strategies to enhance data diversity and quality by stabilizing the denoising score matching training process. 

\textit{CoDi} addresses training challenges with mixed-data types using a dual diffusion model approach. One model handles continuous features and the other manages discrete (categorical) features. Both models use a UNet-based architecture with linear layers instead of traditional convolutional layers, and are trained to condition on each other's outputs. This fixed conditioning setup is designed specifically to handle the interactions between continuous and categorical variables effectively. 

\textit{Tabsyn} applies the concepts of latent diffusion models \citep{rombach2022highresolution, vahdat2021scorebased} to tabular data. First, a VAE encodes mixed-type data into a continuous latent space and then a diffusion model learns this latent distribution. The VAE uses a transformer encoder-decoder to capture feature relationships, while the diffusion model uses an MLP in the reverse denoising process. There are no transformers used in the denoising step, and it does not include a mask conditional method.

Finally, \textit{TabDDPM} manages datasets with mixed data types by using Gaussian diffusion for continuous features and multinomial diffusion for categorical ones. In the preprocessing step, continuous features are scaled using a min-max scaler, and categorical features are one-hot encoded. Then each type of data is sent to its specific diffusion process. After the denoising process, the preprocessing is reversed by scaling back continuous variables, and categorical ones are estimated by first applying a softmax function and then selecting the most likely category.
For classification datasets, \textit{TabDDPM} adopts a class-conditional model consisting in the addition of the condition embedding to the output of the model, while for regression datasets, it integrates target values as an additional feature. It utilizes an MLP architecture as the denosing network optimized with a hybrid objective function that includes mean squared error and KL divergence to predict continuous and categorical data.

Our method is based on \textit{TabDDPM}, following the same logic of having two separate diffusion models for continuous and categorical data. To this end, we augment the model with three key improvements. First, we employ a transformer-based encoder-decoder as the denoising model, which enhances the capability to learn inter-feature interactions for both, condition and masked features. Second, we integrate the conditioning directly into the transformer's attention mechanism rather than simply adding embeddings: this approach reduces learning bias, improving how the interaction between the condition and masked features is modeled. Third, we enable dynamic masking during training, which allows our model to handle varying numbers of visible variables, thus supporting both synthetic data generation and missing data imputation within a single framework. In the following sections, we demonstrate that these contributions lead to improved results across various datasets, generally outperforming \textit{TabDDPM} and other state-of-the-art algorithms.

\section{Background}

\textbf{Diffusion models}, as introduced by \citet{sohl2015deep} and \citet{ho2020denoising}, involve a two-step process: first degrading a data distribution using a forward diffusion process and then restoring its structure through a reverse process. Drawing insights from non-equilibrium statistical physics, these models employ a forward Markov process which converts a  complex unknown data distribution into a simple known distribution (e.g., Gaussian) and vice-versa a generative reverse Markov process that gradually transforms a simple known distribution into a complex data distribution.

More formally, the forward Markov process $q\left(x_{1:T} \middle| x_{0}\right){=}\prod_{t=1}^Tq\left(x_t \middle| x_{t - 1}\right)$ gradually adds noise to an initial sample $x_0$ from the data distribution $q\left(x_0\right)$ sampling noise from the predefined distributions $q\left(x_t \middle| x_{t - 1}\right)$ with variances $\left\lbrace \beta_1, \dots, \beta_T \right\rbrace$. Here $t \, \in \left[ 1 , T \right]$ is the timestep, $T$ is the total number of timesteps used in the forward/reverse diffusion processes and $1{:}T$ means the range of timesteps from $t=1$ to $t=T$.

The reverse diffusion process $p\left(x_{0:T}\right){=}\prod_{t=1}^Tp\left(x_{t-1} \middle| x_{t}\right)$ gradually denoises a latent variable $x_T{\sim}q\left(x_T\right)$ and allows generating new synthetic data. Distributions $p\left(x_{t - 1} \middle| x_t\right)$ are approximated by a neural network with parameters $\theta$. 

In this work, we use the hat notation (e.g., $\hat{p}\left(x_{t - 1} \middle| x_t\right)$) to indicate that a variable is estimated by a neural network model trained with parameters $\theta$. Although these estimations depend on $\theta$, we omit $\theta$ for notational simplicity.
Thus, $\hat{p}\left(x_{t - 1} \middle| x_t\right)$ should be interpreted as $\hat{p}_{\theta}\left(x_{t - 1} \middle| x_t\right)$, where the model parameters $\theta$ influence the predicted value.

The parameters are learned optimizing a variational lower bound (VLB): 
\begin{alignat}{2}
    L_{\text{vlb}} & := \sum_{t=0}^{T} L_t \label{eq:elbo} \\
    L_{0} & := -\log \hat{p}\left(x_0 \middle| x_1 \right) \label{eq:loss0} \\
    L_{t-1} & := \kld{ q \left( x_{t-1} \middle| x_t , x_0 \right)}{\hat{p} \left( x_{t-1} \middle| x_t \right)} \label{eq:losst} \\
    L_{T} & := \kld{ q \left(x_T \middle| x_0 \right)}{p \left( x_T \right)} \label{eq:lossT}
\end{alignat}

The term $q\left( x_{t - 1} \middle| x_t, x_0 \right) $ is the \textit{forward process posterior distribution} conditioned on $x_t$ and on the initial sample $x_0$. $L_{t-1}$ is the Kullback-Leibler divergence between the posterior of the forward process and the estimated reverse diffusion process $\hat{p}\left(x_{t - 1} \middle| x_t\right)$.

\textbf{Gaussian diffusion models} operate in continuous spaces $\left(x_t \in \mathbb{R}^n\right)$ and in this case the aim of the forward Markov process is to convert the complex unknown data distribution into a known Gaussian distribution. This is achieved by defining a forward noising process $q$ that given a data distribution $x_0 \sim q\left(x_0\right)$, produces latents $x_1$ through $x_T$ by adding Gaussian noise at time $t$ with variance $\beta_t \in (0,1)$.
\begin{equation}
\label{eq:gauss_forw}
    \begin{gathered}
        q\left(x_t \middle| x_{t - 1} \right) := \mathcal{N} \left( x_t; \sqrt{1 - \beta_t} x_{t - 1}, \beta_t I \right) \\
        q\left( x_T \right) := \mathcal{N}\left(x_T; 0, I \right) \\
    \end{gathered}
\end{equation}
If we know the exact reverse distribution $q \left( x_{t-1} \middle| x_t \right)$, by sampling from $x_T \sim \mathcal{N} \left( 0 , I \right)$, we can execute the process backward to obtain a sample from $q \left( x_0 \right)$. However, given that $q \left( x_{t-1} \middle| x_t \right)$ is influenced by the complete data distribution, we employ a neural network for its estimation:
\begin{equation}
\hat{p}\left( x_{t - 1} \middle| x_t \right):= \mathcal{N}\left(x_{t - 1}; \hat{\mu}\left(x_t, t\right), \hat{\Sigma}\left(x_t, t\right)\right)  \label{eq:ptheta}
\end{equation}

\citet{ho2020denoising} proposes a simplification of Eq.~\ref{eq:ptheta} by employing a diagonal variance $\hat{\Sigma}\left(x_t, t\right) = \sigma_t I$, where $\sigma_t$ are constants dependent on time. This narrows down the prediction task to $\hat{\mu}\left(x_t, t\right)$. While a direct prediction of this term via a neural network seems the most intuitive solution, another approach could involve predicting $x_0$ and then leveraging earlier equations to determine $\hat{\mu}\left(x_t, t\right)$. Alternatively, it could be inferred by predicting the noise $\epsilon$, as done by \citet{ho2020denoising}. In this work, the authors propose the following parameterization:
\begin{equation}
    \hat{\mu}\left(x_t, t\right) = \frac{1}{\sqrt{\alpha_t}}\left(x_t -\frac{\beta_t}{\sqrt{1 - \bar{\alpha}_t}}\hat{\epsilon}\left(x_t, t\right)\right) \label{eq:gauss_sampling}
\end{equation}
where $\hat{\epsilon}\left( x_t, t \right)$ is the prediction of the noise component $\epsilon$ used in the forward diffusion process between the timesteps $t-1$ and $t$, and $\alpha_t := 1- \beta_t, \ \bar{\alpha}_t := \prod_{i \leq t} \alpha_i$.

The objective Eq.~\ref{eq:elbo} can be finally simplified to the sum of mean-squared errors between $\hat{\epsilon}\left(x_t, t\right)$ and $\epsilon$ over all timesteps $t$:
\begin{equation}\label{eq:mse_simple}
    L_{\text{simple}} = E_{t,x_0,\epsilon}\left[ \left|\left| \epsilon - \hat{\epsilon} \left( x_t, t \right) \right|\right|^2 \right]
\end{equation}

For a detailed derivation of these formulas and a deeper understanding of the methodologies, readers are referred to the original paper by \citet{ho2020denoising, nichol2021improved}.

\textbf{Multinomial diffusion models} \citet{hoogeboom2021argmax} designed the procedures for generating categorical data, where $x_{t} \in \left\lbrace 0, 1 \right\rbrace^{Cl}$ is a one-hot encoded categorical variable with $Cl$ classes. Here, the aim of the forward Markov process is to convert the complex unknown data distribution into a known uniform distribution. The multinomial forward diffusion process $q\left(x_t \middle| x_{t - 1}\right)$ is a categorical distribution that corrupts the data by uniform noise over $Cl$ classes:
\begin{equation}
\label{eq:multi_forw}
    \begin{gathered}
        q \left( x_t \middle| x_{t - 1} \right) := Cat\left(x_t; \left(1 - \beta_t\right)x_{t - 1} + \beta_t / Cl\right) \\
        q\left(x_T\right) := Cat\left(x_T; 1/Cl\right) \\
        q\left(x_t \middle| x_{0}\right) := Cat\left(x_t; \bar{\alpha}_t x_{0} + \left(1 - \bar{\alpha}_t\right)/ Cl\right)
    \end{gathered}
\end{equation}
Intuitively, at each timestep, the model updates the data by introducing a small amount of uniform noise $\beta_t$ across the $Cl$ classes combined with the previous value $x_{t-1}$, weighted by $(1 - \beta_t)$. This process incrementally introduces noise while retaining a significant portion of the prior state, promoting the gradual transition to a uniform distribution. This noise introduction mechanism allows for the derivation of the forward process posterior distribution $q \left( x_{t - 1} \middle| x_t, x_0 \right)$ from the provided equations as follows:
\begin{equation}
    q\left(x_{t - 1} \middle| x_t, x_0\right) = Cat\left(x_{t - 1};  \pi / \sum^{Cl}_{k = 1} \pi_k\right)
    \label{eq:cat_sampling}
\end{equation}
where $\pi = \left[\alpha_t x_t + \left( 1 - \alpha_t \right) / Cl\right] \odot \left[\bar{\alpha}_{t - 1}x_0 + \left( 1 - \bar{\alpha}_{t - 1} \right) / Cl\right]$.

The reverse distribution $\hat{p}\left(x_{t - 1} \middle| x_t\right)$ is parameterized as $q\left(x_{t - 1} \middle| x_t, \hat{x}_0(x_t, t)\right)$, where $\hat{x}_0$ is predicted by a neural network. Specifically, in this approach, instead of estimating directly the noise component $\epsilon$, we predict $x_0$, which is then used to compute the reverse distribution. Then, the model is trained to maximize the VLB Eq.~\ref{eq:elbo}.

\section{\modelname}\label{sect:method}

\textit{\modelname{}} shares foundational principles with the \textit{TabDDPM} approach \citep{kotelnikov2022tabddpm}, but introduces enhancements in both the denoising model and the conditioning mechanism using a transformer encoder-decoder architecture. These improvements enhance the quality of synthetic data and strengthen the conditioning process needed for the reverse diffusion. As a result, the model is capable of generating conditioned synthetic data and performing missing data imputation, generally improving on the performance of other state-of-the-art models for these tasks.

\subsection{Problem definition}
In this work, we address the challenge of modeling tabular datasets for supervised tasks. Our approach is designed to handle both numerical and categorical features, while also effectively dealing with missing values. We focus on tabular datasets for supervised tasks $ D = \left\lbrace \left( x^{c}_{i}, x^{n}_{i}, y_i \right)  \right\rbrace_{i=1}^{N} $,
where $x^{n}_{i} \in \mathbb{R}^{K_{\text{num}}}$ represents the set of numerical features,
$x^{c}_{i} \in \mathbb{Z}^{K_{\text{cat}}}$ represents the set of categorical features, with each categorical feature potentially having a different number of categories, 
and $y_i$ is the target label for row $i$. The dataset contains $N$ rows,
where $K_{\text{num}}$ and $K_{\text{cat}}$ are the number of numerical and categorical features respectively, with the total number of features being $K = K_{\text{num}} + K_{\text{cat}}$.

In our approach, we model numerical features with Gaussian diffusion and categorical features with multinomial diffusion. Each feature is subjected to a distinct forward diffusion procedure, which means that the noise components for each feature are sampled individually.

\textit{\modelname{}} generalizes the approach of \textit{TabDDPM} where the model originally learned $p \left(x_{t-1} \middle| x_t, y \right)$, i.e. the probability distribution of $x_{t-1}$ given $x_t$ and the target $y$.  We extend this by allowing conditioning not only on the target $y$ but also on a subset of input features, aligning with the strategies proposed by~\citet{zheng2022diffusion} and~\citet{tashiro2021csdi}. Specifically, we partition variable $x$ into two subsets:  
$\xmask$ and 
$\xcond$. Here, $\xmask$ contains the features masked and subjected to forward diffusion, while $\xcond$ represents the untouched variable subset that conditions the reverse diffusion. This setup models $ p\left( \xmask_{t-1} \middle| \xmask_t , \xcond , y \right)$, with $\xcond$ and $y$ remaining constant across timesteps $ t $. This approach not only enhances model performance in data generation, but it also enables the possibility of performing data imputation with the same model. 

The reverse diffusion process $p\left(\xmask_{t-1} \middle| \xmask_t, \xcond, y\right)$ is parameterized by a single neural network shown in Figs. \ref{fig:pipeline}, \ref{fig:transformer} and \ref{fig:attention_mechanism}.
This common neural network predicts simultaneously both the amount of noise added for numerical features between steps $t-1$ and $t$ and the distribution of categorical features at time \( t = 0 \). The output dimensionality is \( K_{\text{num}} + \sum_{i=1}^{K_{\text{cat}}} Cl_i \), where \( Cl_i \) denotes the number of classes for the \( i \)-th categorical feature.

\subsection{Model Overview}

\begin{figure}[t]
	\centering 
    \includegraphics[width=0.95\linewidth]{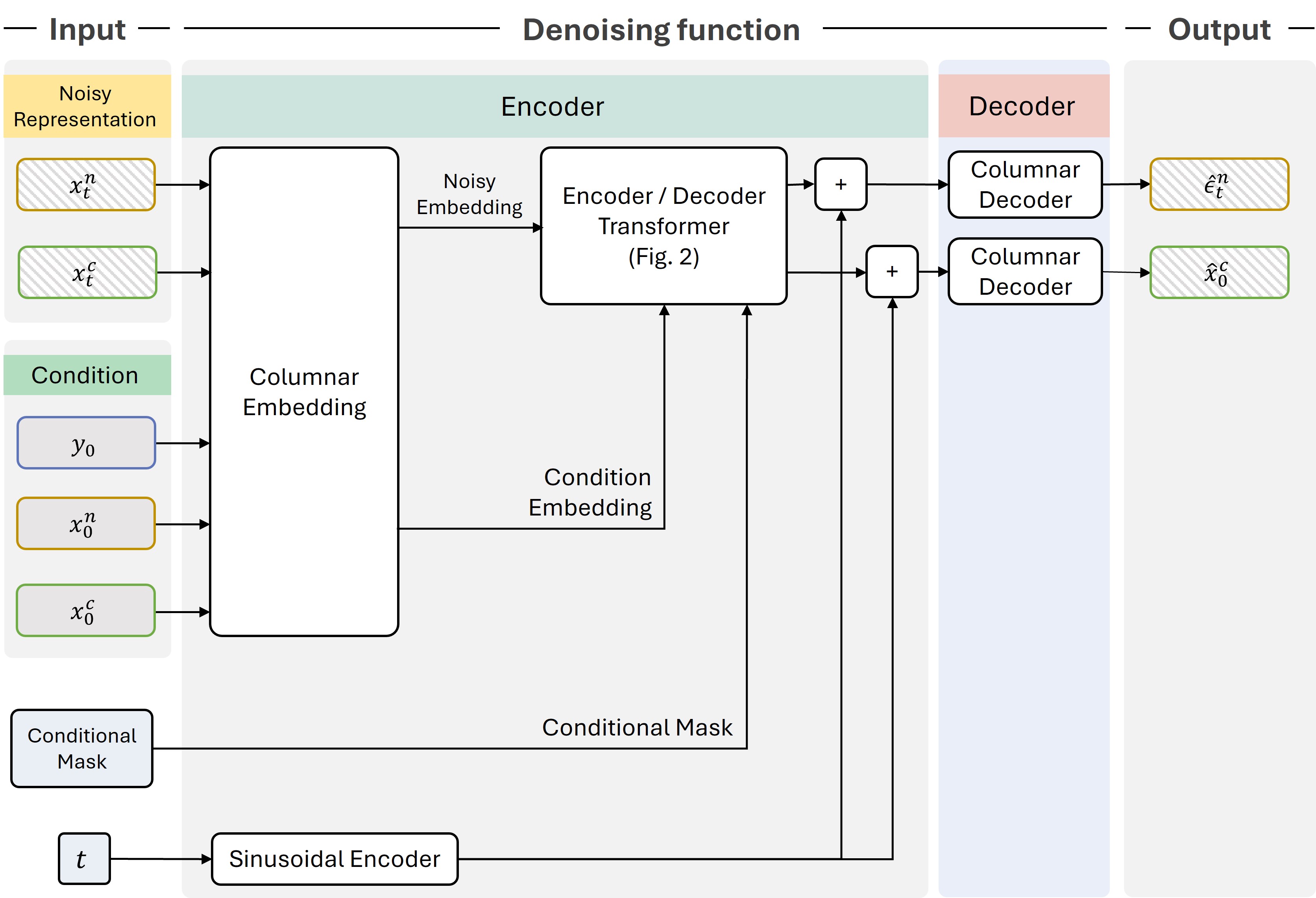}
    \caption{\textit{Denoising function $dn_{\theta}$}. The function takes noisy data representations $x_t^{n}$, $x_t^{c}$, conditioning values $\xcond=  x_0^{n}, x_0^{c}$ and conditioning target $y_0$, a conditional mask $\texttt{Mask}$ and current timestep $t$ as inputs. The \texttt{Columnar Embedding} layer projects numerical and categorical features into a shared latent space, while the \texttt{Encoder-Decoder Transformer} (detailed in Fig. \ref{fig:transformer}) refines this representation by modeling intra- and inter-feature relationships. Outputs are generated through a \texttt{Columnar Decoder}, predicting the estimated noise for numerical features ($\hat{\epsilon}^{n}_{t}$) and logits for categorical features ($\hat{x}^{c}_{0}$).}
    \label{fig:pipeline}
\end{figure}

\modelname{} utilizes an encoder-decoder transformer architecture as the denoising function to refine the representation of noisy data and produce clean samples through reverse diffusion. The workflow is depicted in Figure~\ref{fig:pipeline}, with the following key components: the columnar embedding, the encoder-decoder transformer, and the output decoders for each feature type.

\subsubsection{Denoising Function Architecture}
The denoising function $dn_{\theta}$ is responsible for reconstructing the original data representation from noisy samples. Figures~\ref{fig:pipeline}, \ref{fig:transformer}, and \ref{fig:attention_mechanism} illustrate its detailed workflow and components. The function follows an encoder-decoder model structure and takes as inputs the current noisy representations of numerical and categorical data ($x^{n}_{t}$ and $x^{c}_{t}$) at timestep $t$, a conditional mask (\texttt{Mask}) identifying features involved in the forward diffusion process, the conditioning values ($\xcond$), the target ($y$), and the current timestep ($t$). 

The primary objective of the denoising function is to produce two outputs at each time step: the estimated noise ($\hat{\epsilon}^{n}_{t}$) for numerical features and the logits ($\hat{x}^{c}_{0}$) for categorical features, representing their predicted original distribution. These estimated outputs are then used by Eqs.~\ref{eq:gauss_sampling} and \ref{eq:cat_sampling}, respectively,  to perform the reverse denoising at each timestep $t$ as described in the following sections.

\subsubsection{Columnar Embedding}

To facilitate the learning of the denoising function within the transformer model, a \textit{columnar embedding} projects both numerical and categorical features of $\xmask$ and $\xcond$ into a shared latent space. In the case of categorical features, an embedding layer maps each categorical value into a dense vector representation, whereas numerical features are embedded using a linear transformation followed by a ReLU activation function. The embedding of the target variable \( y \) depends on the supervised task. For regression, \( y \) is treated as a continuous variable, while for classification, it is processed through an embedding layer similar to other categorical features.

\subsubsection{Conditional Encoder-Decoder Transformer}

\begin{figure}[ht!]
	\centering
    \includegraphics[width=0.95\linewidth]{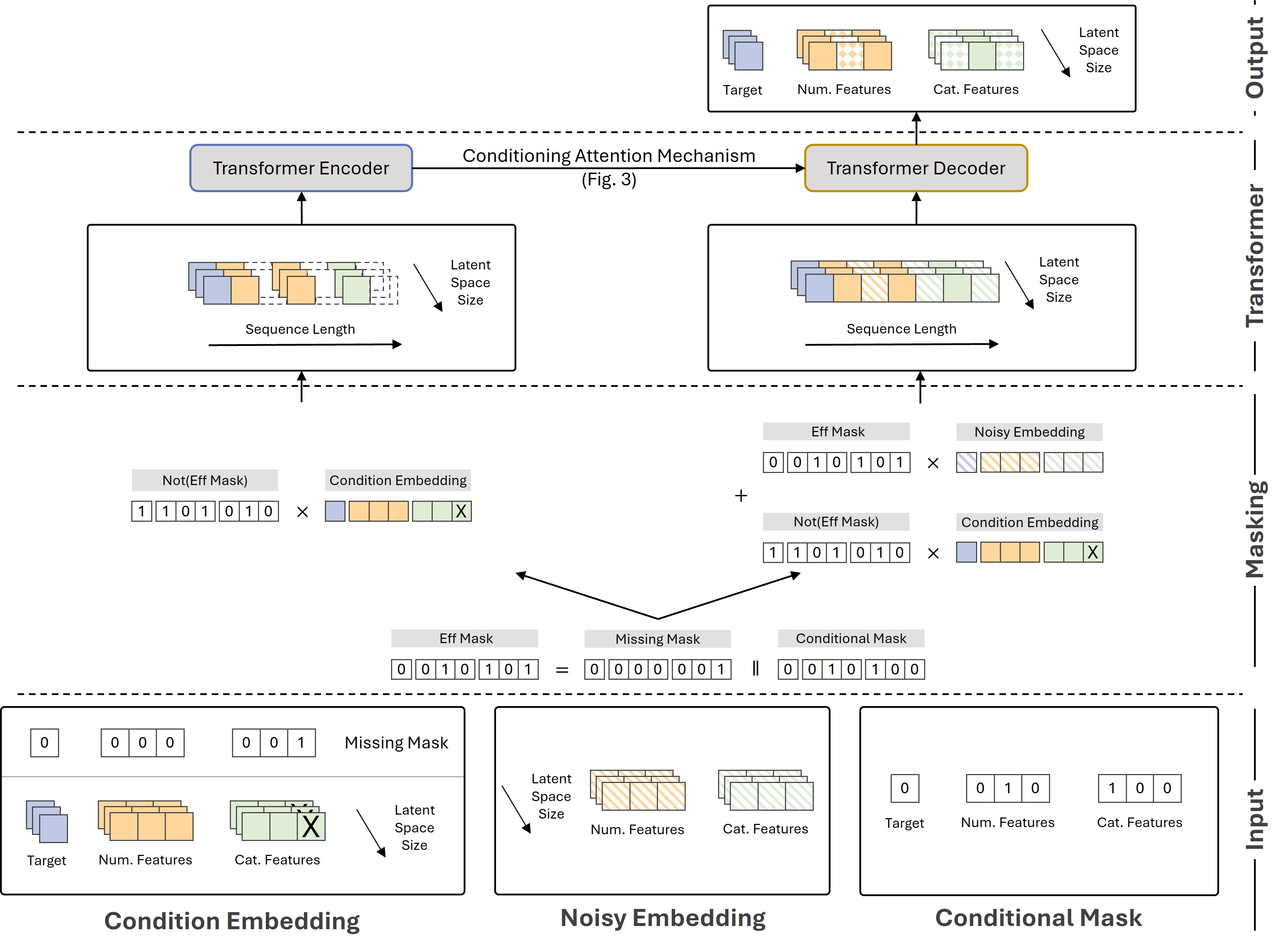}
    \caption{\textit{Conditional Transformer Encoder-Decoder model}. The encoder takes three inputs: noisy data embedding, condition embedding, and a conditional mask (\texttt{Mask}). The
embedding of a conditioning feature is represented by a fully filled square, while the embedding of a noisy feature is depicted as a
dashed-colored square. The effective mask (\texttt{EffMask}) combines missing and conditional masks, allowing the encoder to process conditioning features and target that are unaffected by forward diffusion and without missing values, learning a conditioning context vector for the decoder. 
The decoder then refines the representation of masked features using all available information, including the encoder context, both masked and conditioning features and targets. The final representation of the masked features is depicted as squares filled with colored rhombuses.}
    \label{fig:transformer}
\end{figure}

The workflow, illustrated in Fig.~\ref{fig:transformer}, consists of three inputs, two intermediate stages—masking and transformer—and a final output, arranged from bottom to top. The three inputs are: 
\begin{enumerate*}[label={(\arabic*)}]
\item the columnar embedding of the noisy data, referred to as the noisy embedding, 
\item the columnar embedding of the conditioning data, termed the condition embedding, and 
\item the conditional mask, denoted as $\texttt{Mask}$
\end{enumerate*}. In the conditional mask, features involved in the forward diffusion process are indicated by $\texttt{Mask} = 1$, while features not involved are marked as $\texttt{Mask} = 0$. If the original dataset contains missing values, an additional mask, termed $\texttt{MissingMask}$, is introduced, where $\texttt{MissingMask} = 1$ for missing values, as shown in Fig.~\ref{fig:transformer}.

During the masking phase, an effective mask ($\texttt{EffMask}$) is defined as the logical OR between the conditional mask and the missing value mask, i.e., $\texttt{EffMask} = \texttt{Mask} || \texttt{MissingMask}$. This $\texttt{EffMask}$ is then utilized to generate the input expected by the transformer encoder and decoder.

In the transformer phase, the encoder exclusively processes the conditioning features characterized by $\texttt{EffMask} = 0$, which correspond to features unaffected by the forward diffusion process and those that do not contain missing values. Simultaneously, the decoder processes two distinct types of inputs: 
\begin{enumerate*}[label={(\arabic*)}]
\item a combination of noisy data, which includes features involved in the forward diffusion process or containing missing values ($\texttt{EffMask} = 1$), and conditioning features ($\texttt{EffMask} = 0$), and
\item the output from the encoder
\end{enumerate*}. This setup enables the encoder to provide contextual information from the conditioning features, allowing the decoder to generate refined representations of the features that require denoising. The output of the decoder corresponds to the final output stage, as shown in Fig.~\ref{fig:transformer}.

\begin{figure}[ht!]
	\centering
    \includegraphics[width=3cm]{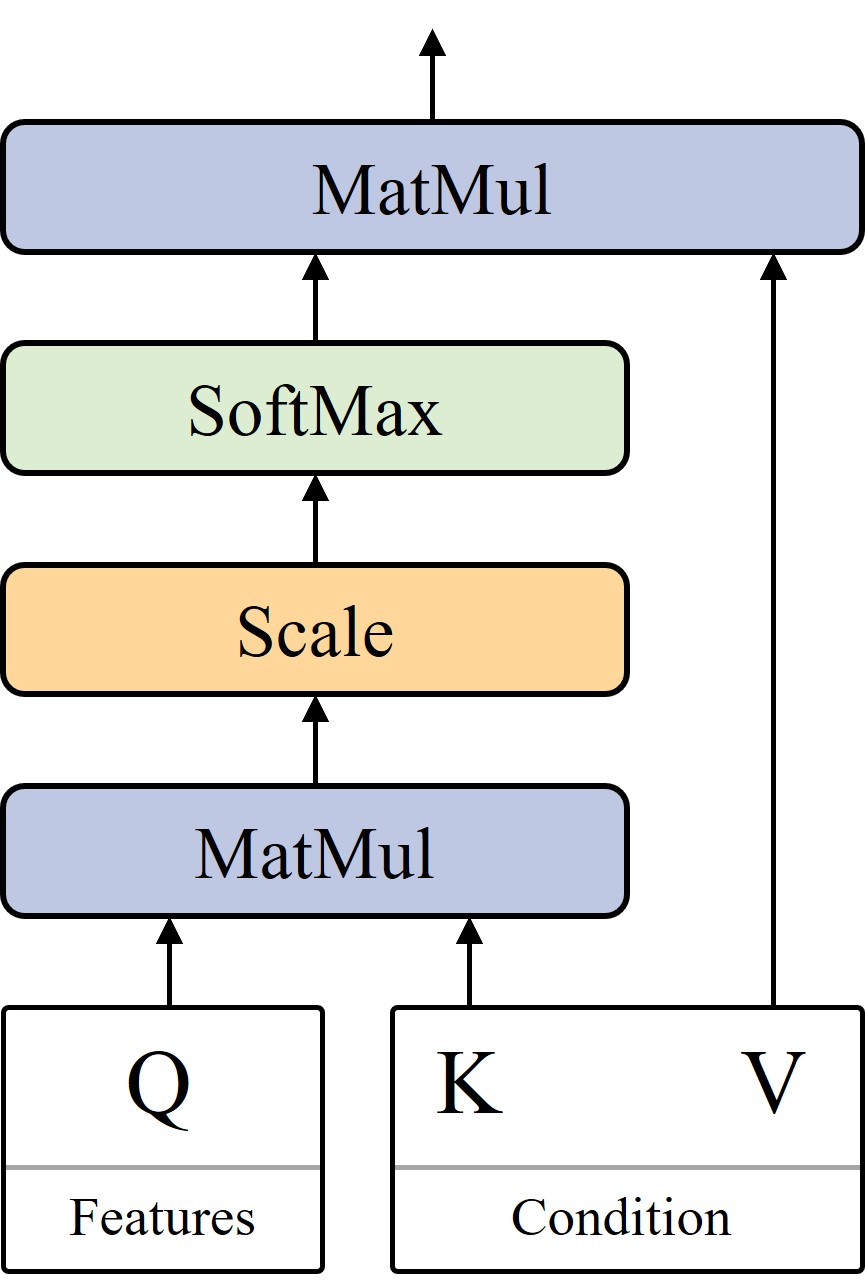}
    \caption{Conditioning Attention Mechanism: The condition embedding produced by the transformer encoder is used in the decoder attention mechanism. More in details, the condition embedding plays the roles of $\bf{K}$ and $\bf{V}$ whereas the feature embedding plays the role of $\bf{Q}$.}
    \label{fig:attention_mechanism}
\end{figure}

\paragraph{Decoder and Attention Mechanism}
The decoder attends to all conditioning features, using the context provided by the encoder. The attention mechanism, depicted in Figure~\ref{fig:attention_mechanism}, incorporates the output of the encoder as the key ($\bf{K}$) and value ($\bf{V}$) vectors, while the features, including the noisy data, serves as the query ($\bf{Q}$) vector, as described in \citet{vaswani2017attention}. This enables the decoder to refine masked feature representations by selectively focusing on interactions between conditioning variables and noisy data.

This attention-based conditioning mechanism is more general and exhibits less learning bias compared to recent approaches in the literature on diffusion models for tabular data, such as \textit{TabDDPM} and \textit{TabCSDI}, where the condition embedding is only added to the masked feature embedding. This advantage holds true even in scenarios where the encoder processes only one variable (the target variable of a supervised task). Here, the transformer encoder functions similarly to an MLP, but its output continues to be used in the decoder's attention mechanism, enhancing overall performance (see discussion on ablation study in Sec.~\ref{sec:Results}).

\subsubsection{Output Denoising Parameters}
Following with the description of the workflow of Fig. \ref{fig:pipeline}, the final latent  representation of the features to denoise is obtained by summing the conditional encoder-decoder transformer output with the timestep embedding, which is derived by projecting the sinusoidal temporal embedding \citep{nichol2021improved, dhariwal2021diffusion} into the transformer embedding dimension, using a linear layer followed by the Mish activation function \citep{misra2020mish}. Last, this representation is decoded to produce the output. Each feature has its own \textit{decoder} consisting of two successive blocks integrating a linear transformation, normalization, and ReLU activation. Depending on the feature type (numerical or categorical), an additional linear layer is appended with either a singular output for numerical features or multiple outputs, corresponding to the number of classes for categorical ones.

\subsection{Sampling or Reverse Diffusion Process} \label{sec:sampling}

The primary goal of the sampling or reverse diffusion process is to convert a set of samples from either Gaussian (for numerical variables) or uniform distributions (for categorical variables) into clean samples that represent a complex real-world distribution.
The sampling proceeds iteratively, moving from timestep $t=T$, where the distribution is simple and noisy, to timestep $t=0$, where the representation becomes clean and realistic.

\begin{algorithm}[h]
\caption{Sampling or Reverse diffusion process}
\label{alg:sampling}
\begin{algorithmic}[1]

\State \textbf{Inputs}: Denoising function $dn_{\theta}$, Conditional Mask $\texttt{Mask}$, Conditioning Value $\xcond$, $y$ and number of Timesteps $T$ 
\State Preprocess the conditioning value $\xcond$, $y$. 
\State Initialize noise:
\State \hskip1.5em Sample initial numerical gaussian noise: $x^{n}_{T} \sim \mathcal{N}\left(x^{n}_T; 0, I \right)$. 
\State \hskip1.5em Sample initial categorical uniform noise: $x^{c}_{T} \sim Cat\left(x^{c}_T; 1/Cl\right)$.
\For{t=T, T-1, ..., 2, 1, 0}
    \State $\hat{\epsilon}^{n}_{t}, \hat{x}^{c}_{0} = dn_{\theta}\left(x^{n}_{t}, \,x^{c}_{t}, \,\texttt{Mask}, \,\xcond, \,y, \,t\right)$.\label{alg:sampling:denoising}
    \State Perform a reverse Gaussian diffusion step using Eq.~\ref{eq:gauss_sampling} to compute $x^{n}_{t-1}$ with $\hat{\epsilon}^{n}_{t}$ and $x^{n}_{t}$.
    \State Perform a reverse multinomial diffusion step using Eq.~\ref{eq:cat_sampling} to compute $x^{c}_{t-1}$ with $\hat{x}^{c}_{0}$ and $x^{c}_{t}$.
\EndFor
\State Postprocess $x^{n}_{0}$ and $x^{c}_{0}$ to reverse the preprocessing from state 2.
\end{algorithmic}
\end{algorithm}

The iterative reverse diffusion process for the generation of one sample is described in Algorithm \ref{alg:sampling}. This algorithm requires four inputs: the denoising function, the conditional mask, the conditioning values, and the number of timesteps. The denoising function, implemented as a neural network, removes noise from the data and is shared across all variables to enhance its capacity to learn complex relationships. The conditional mask distinguishes the conditioning variables, $\xcond$ and $y$, from those involved in the forward diffusion process that require denoising. Variables requiring denoising are marked with $\texttt{Mask} = 1$, while those that do not are marked with $\texttt{Mask} = 0$. The conditioning values are given by $\xcond$ and $y$ at time $t=0$. The number of timesteps dictates how many reverse diffusion steps must be performed.

First, a consistent preprocessing procedure is applied to the conditioning values, using the Gaussian quantile transformation from the scikit-learn library \citep{scikit-learn} for numerical features and ordinal encoding for categorical ones. Then, at time $t=T$, the process initiates by sampling the initial noise for each dataset feature separately: Gaussian noise for numerical features and uniform noise for categorical features. The iterative denoising process then proceeds from $t=T$ to $t=0$. Each iteration involves three key actions: first, the denoising function $dn_{\theta}$ estimates $\hat{\epsilon}^{n}_{t}$ for continuous features and $\hat{x}^{c}_{0}$ for categorical features. Next, a reverse Gaussian step and a reverse multinomial step are performed using Eqs.~\ref{eq:gauss_sampling} and \ref{eq:cat_sampling} to estimate $x^{n}_{t-1}$ and $x^{c}_{t-1}$, respectively. These actions are repeated iteratively until the final outputs $x^{n}_{0}$ and $x^{c}_{0}$ are reached. Finally, a postprocessing step is applied to the final outputs $x^{n}_{0}$ and $x^{c}_{0}$ to reverse the initial preprocessing. It is important to note that the preprocessing of conditioning values and postprocessing of outputs enhances the learning rate without impacting the quality of the final results. We tested various normalization methods for numerical features, such as Standard Scaler and Min-Max Scaler from scikit-learn, and found the results to be statistically equivalent.

\subsection{Training process}

The training process for the denoising function is outlined in Alg. \ref{alg:training}.
The primary aim is to train the neural network $dn_{\theta}$ used in the state \ref{alg:sampling:denoising} of Alg. \ref{alg:sampling}. This network is designed to leverage noisy representations of numerical and categorical features $x^{n}_{t}, x^{c}_{t}$, along with the conditional mask $\texttt{Mask}$, conditioning values, and current timestep $t$, to estimate both $\hat{\epsilon}^{n}_{t}$ and $\hat{x}^{c}_{0}$. For numerical features, $\hat{\epsilon}^{n}_{t}$ represents the estimated noise introduced during the forward diffusion process between timesteps $t-1$ and $t$. For categorical features, $\hat{x}^{c}_{0}$ serves as an estimate of the original categorical feature a $t=0$.

\begin{algorithm}
\caption{Training process}
\label{alg:training}
\begin{algorithmic}[1]
\State \textbf{Inputs}: Denoising function $dn_{\theta}$, number of Timesteps $T$, original data $x^{n}_{0}, x^{c}_{0}$
\State Initialize the weights $\theta$ of the Denoising function $dn_{\theta}$.
\State Preprocess the original data $x^{n}_{0}, x^{c}_{0}$
\For{iteration=1, 2, ..., N}
    \State Sample a batch of original features, $x^{n}_{0},\,\,x^{c}_{0}$. 
    \State Sample a batch of timesteps $t$ uniformly distributed within the interval $\left[ 0, T \right]$.
    \For{each numerical features $x^{n}_{0}$}
        \State Use the forward Gaussian diffusion process of Eq.~\ref{eq:gauss_forw} (with reparametrization trick) to obtain $x^{j_n}_{t}$ and $\epsilon^{n}_{t}$. 
    \EndFor
    \For{each categorical feature $x^{c}_{0}$}
        \State Use the forward multinomial diffusion process of Eq.~\ref{eq:multi_forw} to obtain $x^{c}_{t}$.
    \EndFor
    \State Generate the new conditional $\texttt{Mask}$
    \State Compute the denoising function outputs: $\hat{\epsilon}^{n}_{t}, \hat{x}^{c}_{0} = dn_{\theta} \left( x^{n}_{t}, x^{c}_{t}, \texttt{Mask}, x^{n}_{0}, x^{c}_{0}, t \right)$.
     \State Use Eq.~\ref{eq:mse_simple} to compute $L_{simple}\left(\texttt{Mask}\right)$ for numerical features with $\texttt{Mask} = 1$; using $\epsilon^{n}_{t}$ and $\hat{\epsilon}^{n}_{t}$
    \State Use Eq.~\ref{eq:elbo} to compute $L_t\left( \texttt{Mask} \right)$ for categorical features with $\texttt{Mask} = 1$; using $x^{c}_{0}$, $ \hat{x}^{c}_{0}$ and $x^{c}_{t}$
    \State Compute the total loss function of Eq.~\ref{eq:loss_tot}.
    \State Minimize the total loss function and update the denoising function weights.
\EndFor
\end{algorithmic}
\end{algorithm}

The first steps involve initializing the model weights and preprocessing the original data as discussed in the previous section, followed by an iterative learning process. In each iteration, a batch of original features $x^{n}_{0},\,\,x^{c}_{0}$ is sampled from the original dataset.  For each element in the batch, a timestep $t$ is then sampled uniformly from the interval $\left[ 0, T \right]$. Using the forward Gaussian diffusion process, targets for numerical features are computed. Specifically, for each numerical feature, the reparametrization trick is applied to Eq.~\ref{eq:gauss_forw} to compute $x^{n}_{t}$, that is the noisy representation of the numerical feature at time $t$, and $\epsilon^{n}_{t}$, the corresponding noise. Similarly, the forward multinomial diffusion process of Eq.~\ref{eq:multi_forw} is used to obtain $x^{c}_{t}$, the noisy representation of the categorical feature a time $t$.  Next, a new mask is dynamically generated to select a subset of features that will serve as the conditioning input. With all the all the $dn_{\theta}$ inputs ($x^{n}_{t}$, $x^{c}_{t}$, $\texttt{Mask}$, $x^{n}_{0}$, $x^{c}_{0}$, $t$) and targets ($\epsilon^{n}_{t}$ and $x^{c}_{0}$) in place, the loss function is then calculated. 

The model is trained by minimizing the following total loss function: 
\begin{equation}
\label{eq:loss_tot}
    L^{MTabGen}_t = \dfrac{L^{simple}_t(\texttt{Mask})}{K_{num}} + \dfrac{\sum_{i \leq K_{cat}} L^{i}_t(\texttt{Mask})/Cl_i}{K_{cat}} 
\end{equation}
The total loss function $L^{MTabGen}_t$ includes two main components:
\begin{itemize}
    \item Loss for \textbf{numerical features}: $L^{simple}_t$ (defined in Eq.~\ref{eq:mse_simple}) computes the \textit{mean squared error} between the true noise $\epsilon^{n}_{t}$ and the predicted noise $\hat{\epsilon}^{n}_{t}$ introduced in the forward Gaussian diffusion process between the steps $t-1$ and $t$. 
    \item Loss for \textbf{categorical features}:
    $L^{i}_t$ (defined in Eq.~\ref{eq:elbo}) calculates the \textit{Kullback-Leibler} divergence between the posterior of the forward process $q\left( x_{t-1}^{c} \middle| x_{t}^{c}, x_{0}^{c}\right)$ and parametrized reverse diffusion process $\hat{p}\left( x_{t-1}^{c} \middle| x_{t}^{c}\right)$. Following the approach in \citep{hoogeboom2021argmax}, the $\hat{p}\left( x_{t-1}^{c} \middle| x_{t}^{c} \right)$ is approximated by $q\left( x_{t-1}^{c} \middle| x_{t}^{c}, \hat{x}^{c}_{0}\right)$ using Eq.~\ref{eq:cat_sampling}.
\end{itemize}
    
 $L^{simple}_t(\texttt{Mask})$ and $L^{i}_t(\texttt{Mask})$ means that the loss functions are computed taking into account only the prediction error on variables affected by the forward diffusion process (i.e. $\texttt{Mask}=1$). Here, $Cl_i$ is the number of classes of the $i$-th categorical variable. 

\subsection{Dynamic Conditioning}
A key feature of the proposed solution is that the split between $\xmask$ and $\xcond$ does not have to be fixed for every row $i$ in the dataset. The transformer encoder can manage mask with an arbitrary number of zeros/ones, so we can dynamically alter the split between $\xmask$ and $\xcond$ by just producing a new mask. In the extreme scenario, we can generate a new mask $\texttt{Mask}_i$ for each row $i$. During training, the number of ones in  $\texttt{Mask}_i$ (i.e. the number of features to be included in the forward diffusion process) is uniformly sampled from the interval $[1, K_{num} + K_{cat}]$. A model that has been trained in this manner can then be used for both tasks, that is, generation of synthetic data ($\texttt{Mask}_i = 1$ for all the $K_{num} + K_{cat}$ features and for any dataset index $i$) and imputation of missing values (for each $i$, $\texttt{Mask}_i = 1$ for the feature to impute). As discussed, when the original dataset contains missing values,  a new \texttt{MissingMask} is introduced and combined with the conditional \texttt{Mask} to remove any missing value from the condition. The \texttt{MissingMask} is fixed and constant during the training phase. This setup allows for more flexible conditioning scenarios. 

Specifically:
\begin{itemize}
    \item When $\xcond = \emptyset$, $p\left(\xmask_{t-1} \middle| \xmask_{t}, \xcond, y\right) = p\left(x_{t-1} \middle| x_t, y\right)$ our model aligns with \textit{TabDDPM}, generating synthetic data influenced by the target distribution.
    \item When $\bar{x}^M \neq \emptyset$, the model can generate synthetic data based on the target distribution and either a fixed or dynamic subset of features. Conditioning on a fixed subset introduces advantages in settings where certain variables are readily accessible, whereas others are difficult to obtain due to challenges like cost constraints. In such cases, the scarce data can be synthetically produced using the known variables. Conversely, when conditioning on a dynamic subset of features, the model effectively addresses the challenge of imputing gaps within a dataset.
\end{itemize}

\section{Experiments\label{sec:Experiments}}

\subsection{Data} 
Below we introduce the benchmark datasets used in the performance evaluation of our model. The statistics are summarized in Table~\ref{tab:datasets}. 

\begin{itemize}

\item  \textit{HELOC} \cite{heloc}: Home Equity Line of Credit (HELOC)
provided by FICO (a data analytics company), contains
anonymized credit applications of HELOC credit lines. The
dataset contains 21 numerical and two categorical features
characterizing the applicant to the HELOC credit line. The task is a binary classification and the goal is to predict whether the applicant will make timely payments over a two-year period.

\item  \textit{Churn Modelling} \cite{churn_modelling}: This dataset consists of six numerical and four categorical features about bank customers. The binary classification task involves predicting whether or not the customer closed their account.

\item  \textit{Gas Concentrations} \cite{vergara2012chemical}: The dataset contains measurements from 16 chemical sensors exposed to six gases at different concentration levels. It contains $13.9$M of rows and $129$ continuous features and the classification task is to determine which is the gas generating the data.

\item  \textit{California Housing} \cite{californiahousing}: The information refers to the
houses located in a certain California district, as well as some basic statistics about them based on 1990 census data. This is a regression task about forecasting the price of a property.

\item  \textit{House Sales King Country} \cite{house_sales}: Similar to the California Housing case, this regression task involves estimating property prices in the King County region for sales between May 2014 and May 2015. The original dataset included 14 numerical features, four categorical features, and one date feature. During pre-processing, the date feature was transformed into two categorical variables: month and year.

\item  \textit{Adult Incoming} \cite{adult}: Personal details such as age, gender
or education level are used to predict whether an individual would earn more or less than 50K$\$$ per year.

\item  \textit{Otto Group} \cite{otto}: This dataset, provided by the Otto Group (an e-commerce company), contains $61.9$K of rows and $93$ continuous product attributes. The task is a multi-class classification problem with nine categories, aiming to determine the category to which each product belongs.

\item  \textit{Cardiovascular Disease} \cite{cardiovascular_disease}:
The existence or absence of cardiovascular disease must be predicted based on factual information, medical examination results, and information provided by the patient. The dataset consists of seven numerical and four categorical features.

\item  \textit{Insurance} \cite{insurance_company}:
Customer variables and past payment data are used to solve a binary task: determining whether the customer will pay on time. The dataset has eight numerical and two categorical features. 

\item  \textit{Forest Cover Type} \cite{covertype}: In this multi-class classification task with seven categories, cartographic variables are used
to predict the forest cover type. The first eight features of the dataset are continuous, whereas the last two are categorical with 4 and 40 levels respectively.
\end{itemize}

\begin{table}
\caption{Tabular benchmark properties.}
\label{tab:datasets}
\begin{tabular}{lcccc}
    \toprule
    Dataset & Rows & Num. Feats & Cat. Feats & Task \\ 
    \midrule
    HELOC & 9871 & 21 & 2 & Binary \\
    Churn & 10000 & 6 & 4 & Binary \\
    Gas Concentrations & 13910 & 129 & 0 & Multi-Class ($6$) \\
    Cal. Hous. & 20640 & 8 & 0 & Regression \\
    House Sales & 21613 & 14 & 2
    & Regression \\
    Adult Inc.& 32561 & 6 & 8 & Binary \\
    Otto Group & 61900 & 93 & 0 & Multi-Class ($9$) \\
    Cardio & 70000 & 7 & 4 & Binary \\
    Insurance & 79900 & 8 & 2 & Binary \\
    Forest Cov. & 581 K & 10 & 2
    & Multi-Class (7) \\
    \bottomrule
\end{tabular}
\end{table}

\subsection{Baselines}
For the synthetic data generation task, we consider the following state-of-the-art baselines drawn from representative generative modeling paradigms: VAE, GAN and Diffusion Models:
\begin{itemize}
    \item \textit{TabDDPM} \citep{kotelnikov2022tabddpm}: State-of-the-art diffusion model for tabular data generation and model in which we have premised the proposed approach.

    \item \textit{Tabsyn}\footnote{GitHub: \url{https://github.com/amazon-science/tabsyn}} \citep{zhang2024mixedtype}: 
    Recent state-of-the-art tabular generative model that integrates a diffusion model into the continuous latent space projected by a VAE.

    \item \textit{CoDi}\footnote{GitHub: \url{https://github.com/ChaejeongLee/CoDi}} \citep{lee2023codi}: A diffusion model for tabular data generation. The \textit{StaSy} and \textit{CoDi} models are from the same team. In \citep{lee2023codi}, the authors show that \textit{CoDi} consistently outperforms \textit{StaSY}. Therefore, we only include \textit{CoDi} in our evaluation.
    
    \item \textit{TVAE}\footnote{\label{sdv_ref} We use the implementation provided by \url{https://sdv.dev/}} \citep{xu2019modeling}: A variational autoencoder adapted for mixed-type tabular data.
    
    \item \textit{CTGAN}\footref{sdv_ref} \citep{xu2019modeling}: A conditional GAN for synthetic tabular data generator.
\end{itemize}

For the missing data imputation task, the following state-of-the-art baselines have been considered:
\begin{itemize}
    \item \textit{missForest}\footnote{\label{hyperimpute_ref} We use the implementation provided by \url{https://github.com/vanderschaarlab/hyperimpute}} \citep{missForest}: Iterative method based on random forests to predict and fill in missing values.
    \item \textit{GAIN}\footref{hyperimpute_ref}  \citep{yoon2018gain}: GAN model for tabular missing value imputation.
    \item \textit{HyperImpute}\footref{hyperimpute_ref} \citep{jarrett2022hyperimpute}: Iterative imputation algorithm using both regression and classification methods based on linear models, trees, XGBoost, CatBoost and neural nets.
    \item \textit{Miracle}\footref{hyperimpute_ref} \citep{kyono2021miracle}: Missing imputation algorithm using a causal deep learning approach.
    \item \textit{TabCSDI}\footnote{GitHub: \url{https://github.com/pfnet-research/TabCSDI}} \citep{zheng2022diffusion}: State-of-the-art diffusion model for missing data imputation.
\end{itemize}

\subsection{Metrics\label{sec:experiments:metrics}}
We evaluate the generative models on three different dimensions: 
\begin{enumerate*}[label={(\arabic*)}]
\item ML efficiency, 
\item statistical similarity and
\item privacy risk.
\end{enumerate*}

\subsubsection{Machine Learning efficiency} \label{sec:ML_eff} 

The \textit{Machine Learning efficiency} measures the performance degradation of classification or regression models trained on synthetic data, and then tested on real data. The basic idea is to use a ML discriminative model to evaluate the utility of synthetic data provided by a generative model. As demonstrated by \citet{kotelnikov2022tabddpm}, a strong ML model allows to obtain more stable and consistent conclusions on the performances of the generative model. Based on this intuition, we consider 4 different ML models: \textit{XGBoost} \citep{chen2016xgboost}, \textit{CatBoost} \citep{prokhorenkova2018catboost}, \textit{LightGBM} \citep{ke2017lightgbm} and \textit{MLP}. We introduce an initial fine-tuning step, during which we derive the best hyperparameter configuration using Bayesian optimization and Optuna library \citep{akiba2019optuna}. Specifically, we perform 100 iterations to fine-tune the model's (\textit{XGBoost}, \textit{CatBoost}, \textit{LightGBM} and \textit{MLP}) hyperparameters on each dataset's real data within the benchmark. Every hyperparameter configuration for ML model is cross-validated, using a five-fold split. The complete hyperparameter search space is shown in Appendix~\ref{sec:hyperparameter} Table~\ref{tab:optuna_search_space}.

Once the discriminative model has been optimized for each dataset, the generative model is further cross-validated using a five-fold split, by implementing the following procedure. For each fold, the real data is split into three subsets. The main purpose of the first subset is to train the generative model. The resulting model generates a synthetic dataset conditioned on the second subset. The synthetic dataset is then used to train the discriminative model. The so-obtained ML model is finally tested on the third held-out subset, which has not been used in training any of the models. The procedure is repeated for each fold, and the obtained metric mean is used as a final measure to compute the generative model ML efficiency.

\subsubsection{Statistical similarity} \label{sec:Stat_sim} 

The comparison between synthetic and real data accounts for both individual and joint feature distributions. Adopting the approach proposed by \citet{zhao2021ctabgan}, we employ Wasserstein \cite{wang2021two} and Jensen-Shannon distances \cite{lin1991divergence} to analyze numerical and categorical distributions. In addition, we use the square difference between pair-wise correlation matrix to evaluate the preservation of feature interactions in synthetic datasets. Specifically, the Pearson correlation coefficient measures correlations between numerical features, the Theil uncertainty coefficient measures correlations between categorical features, and the correlation ratio evaluates interactions between numerical and categorical features.

\subsubsection{Privacy Risk} \label{sec:Priv_Risk} 

The \textit{Privacy Risk} is evaluated using the Distance to Closest Record (DCR), i.e. the Euclidean distance between any synthetic record and its closest corresponding real neighbour. Ideally, the higher the DCR the lesser the risk of privacy breach. It is important to note that out-of-distribution data, i.e. random noise, will also produce high DCR. Therefore, to maintain ecological validity, the DCR metric needs to be evaluated jointly with the ML efficiency metric.

\section{Results\label{sec:Results}}

\subsection{Machine Learning efficiency}

\subsubsection{Synthetic data generation} \label{sec:syn_data_gen}

In this task, we evaluate the performance of our generative model in producing high-quality synthetic data, conditioned exclusively by the supervised target $y$. To this end, we consider two variants of our model:
\begin{enumerate}[leftmargin=15pt]
    \item \textit{\modelname{}~I}: This variant consistently includes all dataset features in the diffusion process and was specifically designed for the synthetic data generation task.
    \item \textit{\modelname{}~II}: During training, this variant dynamically selects which features are incorporated in the diffusion process, making it versatile for both imputing missing data and generating complete synthetic datasets.
\end{enumerate}

\begin{table}
\caption{\textit{Machine Learning efficiency}. Classification tasks use F1-score, and regression tasks use MSE, indicated by up/down arrows for maximization/minimization of the metric. Cross-validation mean and standard deviation are shown for each dataset-model pair. Best and second-best results are highlighted in bold and underline, respectively. \textbf{Baseline} column shows average performance of ML models trained on real data, while other columns reflect average performance of ML models trained on synthetic data from specified models. All models are tested on real data.}
\label{tab:ml_res}
\resizebox{0.95\linewidth}{!}{
\begin{tabular}{l|c|ccccc|cc}
    \toprule
    Dataset & Baseline & TVAE & CTGAN & CoDi & Tabsyn & TabDDPM & \modelname{}~I  & \modelname{}~II \\ 
    \midrule
    HELOC $\uparrow$        & $83.69 \pm 0.04$  & $79.41 \pm 0.05$  & $77.53 \pm 0.06$  & $75.82 \pm 0.07$ & $79.24 \pm 0.05$ & $76.69 \pm 0.10$  & $\mathbf{82.91\pm 0.07}$   & $\underline{82.71\pm 0.08}$ \\
    Churn $\uparrow$        & $85.25 \pm 0.05$  & $81.67 \pm 0.08$  & $79.31 \pm 0.06$  & $82.77 \pm 0.15$ & $\mathbf{84.60 \pm 0.06}$ & $83.62 \pm 0.15$  & $\underline{84.43\pm 0.03}$   & $83.91\pm 0.04$ \\
    Gas $\uparrow$          & $99.47 \pm 0.09$  & $94.55 \pm 0.07$  & $62.04 \pm 0.07$  & $65.41 \pm 0.04$ & $\underline{98.66 \pm 0.04}$ &  $65.51 \pm 0.06$  & $\mathbf{98.80 \pm 0.07}$   & $98.60 \pm 0.06$ \\
    Cal. Hous. $\downarrow$ & $0.161 \pm 0.001$ & $0.316 \pm 0.003$ & $0.488 \pm 0.004$ & $0.290 \pm 0.003$ & $0.256 \pm 0.002$ &  $0.272 \pm 0.002$ & $\mathbf{0.224 \pm 0.001}$ & $\underline{0.227 \pm 0.001}$ \\
    House Sales $\downarrow$& $0.101 \pm 0.001$ & $0.209 \pm 0.001$ & $0.335 \pm 0.001$ & $0.159 \pm 0.001$ & $0.148 \pm 0.001$ & $\underline{0.145 \pm 0.001}$ & $\mathbf{0.121 \pm 0.001}$ & $\underline{0.145 \pm 0.001}$ \\
    Adult Inc. $\uparrow$   & $86.95 \pm 0.04$  & $84.34 \pm 0.07$  & $83.64 \pm 0.08$  & $84.43 \pm 0.08$ & $84.70 \pm 0.06$ &  $84.86 \pm 0.07$  & $\mathbf{85.30 \pm 0.09}$  & $\underline{85.15 \pm 0.07}$ \\
    Otto $\uparrow$         & $81.50 \pm 0.06$    & $63.87 \pm 0.05$  & $50.48 \pm 0.05$  & $63.21 \pm 0.06$ & $67.14 \pm 0.08$ &  $63.34 \pm 0.08$  & $\mathbf{73.04 \pm 0.07}$  & $\underline{72.98 \pm 0.06}$ \\
    Cardio $\uparrow$       & $73.54 \pm 0.05$  & $72.47 \pm 0.08$  & $71.81 \pm 0.06$  & $72.34 \pm 0.10$ & $\underline{72.90 \pm 0.11}$&  $72.88 \pm 0.14$  & $\mathbf{72.97 \pm 0.08}$  & $72.67 \pm 0.12$ \\
    Insurance $\uparrow$    & $92.00\pm 0.03$  & $92.63 \pm 0.11$  & $92.56 \pm 0.05$  & $92.03 \pm 0.07$ & $92.20 \pm 0.05$ &  $92.21 \pm 0.04$  & $\mathbf{92.77 \pm 0.11}$  & $\underline{92.64 \pm 0.07}$ \\
    Forest Cov. $\uparrow$  & $96.42 \pm 0.06$  & $70.36 \pm 0.04$  & $65.65 \pm 0.07$  & $74.64 \pm 0.09$ & $74.83 \pm 0.10$ &  $82.08 \pm 0.07$  & $\mathbf{85.61 \pm 0.04}$  & $\underline{84.32 \pm 0.09}$ \\
    \midrule
    \multicolumn{2}{c|}{Average Rank} & $4.9$ & $6.5$ & $5.7$ & $3.3$ & $4.0$ & $\mathbf{1.1}$ & $\underline{2.4}$ \\
    \bottomrule
\end{tabular}}
\end{table}

The results shown in Table~\ref{tab:ml_res} indicate that \textit{\modelname{}~II} demonstrates competitive performance compared to existing state-of-the-art methods like \textit{TabDDPM}, \textit{Tabsyn}, \textit{CoDi}, \textit{TVAE}, or \textit{CTGAN} in the synthetic data generation task, while showing moderate but  statistically significant\footref{statistically_significant} improvements in most of the datasets.
However, the specialized \textit{\modelname{}~I} achieves the best performance across the evaluated tasks.
The key outcomes of our experiments are as follows: 
\begin{enumerate*}[label={(\arabic*)}]
\item \textit{TVAE} produces better results than \textit{CTGAN}.
\item Approaches based on Diffusion Models outperform \textit{TVAE} and \textit{CTGAN} on average.
\item Our two proposed models show overall improved performance over \textit{TabDDPM}, \textit{Tabsyn} and \textit{CoDi}, with statistically significant improvements in specific datasets\footnote{\label{statistically_significant} Following the recommendation of \cite{Rainio2024Evaluation}, we applied the Wilcoxon signed-rank test to compare, for each dataset, the results obtained by the best-performing MTabGen model against the best-performing baseline. In 6 out of 10 cases-specifically, HELOC, California Housing, House Sales, Adult Income, Otto, and Forest Cover Type, MTabGen demonstrated a statistically significant improvement over the strongest baseline, with a p-value $< 0.01$.}. In the remaining cases, the results were on par with the baselines, without a statistically significant difference.
\item Our model tends to outperform the baselines in datasets with a large number of features, such as the \textit{Gas Concentrations} and \textit{Otto Group} datasets, although this trend is less consistent with \textit{Tabsyn}. Additionally, it is worth noting that the ML efficiency results align with those reported in the \textit{Tabsyn} paper, where \textit{Tabsyn} generally surpasses \textit{TabDDPM} and \textit{CoDi}, with \textit{TVAE} also showing stronger performance than \textit{CoDi}.

\end{enumerate*}

The results presented in Table~\ref{tab:ml_res} are obtained after applying Bayesian optimisation for each generative model, using the Optuna library over $100$ trials, and evaluating performance with the cross-validated ML efficiency metric defined in Section \ref{sec:experiments:metrics} as the objective. A similar hyperparameter optimization procedure was applied to all baseline models, considering the parameters specific to each method, enabling fair comparison with \textit{\modelname{}}. The specific hyperparameter search space for each model is shown in Table~\ref{tab:model_search_space}, Appendix~\ref{sec:hyperparameter}.

\subsubsection{Ablation study}

We conduct an ablation study to evaluate the contributions of the encoder-decoder transformer and dynamic conditioning to our model's performance and usability. The encoder-decoder transformer enhances performance for two primary reasons:
\begin{itemize}
    \item \textit{Enhanced learning of inter-feature interactions within condition (encoder) and masked features (decoder).} Transformer layers allow for better learning of inter-feature interactions compared to MLPs. This is primarily due to the attention mechanism, where the new embedding of a feature $x_i$ is computed by a linear combination of the embeddings of all features $\left\lbrace x_j \right\rbrace$. The weight of feature $x_j$ depends on the current values of $x_i$ and $x_j$. This mechanism is more flexible than in MLP case, where the contribution of feature $x_j$ to a neuron in the next layer is fixed and does not depend on its current value.
    \item \textit{Conditioning attention mechanism.} Our model, \textit{\modelname{}}, uses an encoder-decoder transformer architecture. The encoder learns latent representations of unmasked features for conditioning, while the decoder focuses on learning latent representations of masked or noisy features. By incorporating conditioning within the attention mechanism of the transformer decoder, we reduce learning bias compared to conventional methods like those used in \textit{TabDDPM}, which simply sum the latent representations of conditions and noisy features.
\end{itemize}

To test the impact of these hypotheses, we modified the original implementation of \textit{TabDDPM} by replacing the MLP denoising model with a transformer encoder, while retaining the same conditioning mechanism (i.e., summing the condition embedding and feature embedding). We call this new implementation \textit{TabDDPM-Transf}. As shown in the first two columns of Table~\ref{tab:ablation}, our transformer-enhanced \textit{TabDDPM-Transf} consistently outperforms the standard \textit{TabDDPM}. The impact of the conditioning attention mechanism is further demonstrated by the comparison of \textit{TabDDPM-Transf} and \textit{\modelname{}~I} in Table~\ref{tab:ablation}.

\begin{table}
\caption{\textit{Ablation study} in terms of \textit{Machine Learning efficiency}. Classification tasks use F1-score, and regression tasks use MSE, indicated by up/down arrows for maximization/minimization of the metric. Cross-validation mean and standard deviation are shown for each dataset-model pair. Best and second-best results are highlighted in bold and underline, respectively. \textbf{Baseline} column shows average performance of ML models trained on real data, while other columns reflect average performance of ML models trained on synthetic data from specified models. All models are tested on real data.}
\label{tab:ablation}
\begin{tabular}{l|c|ccc}
    \toprule
    Dataset & Baseline & TabDDPM & TabDDPM-Transf  & \modelname{}~I  \\ 
    \midrule
    HELOC $\uparrow$        & $83.69 \pm 0.04$  & $76.69 \pm 0.10$  & $\underline{81.15\pm 0.08}$ & $\mathbf{82.91\pm 0.07}$ \\
    Churn $\uparrow$        & $85.25 \pm 0.05$  & $83.62 \pm 0.15$  & $\underline{83.62\pm 0.05}$ & $\mathbf{84.43\pm 0.03}$ \\
    Gas $\uparrow$          & $99.47 \pm 0.09$  & $65.51 \pm 0.06$  & $\underline{93.35 \pm 0.09}$ & $\mathbf{98.80 \pm 0.09}$  \\
    Cal. Hous. $\downarrow$ & $0.161 \pm 0.001$ & $0.272 \pm 0.001$ & $\underline{0.232 \pm 0.001}$ & $\mathbf{0.224 \pm 0.001}$ \\
    House Sales $\downarrow$& $0.101 \pm 0.001$ & $0.145 \pm 0.001$ & $\underline{0.128 \pm 0.001}$ & $\mathbf{0.121 \pm 0.001}$ \\
    Adult Inc. $\uparrow$   & $86.95 \pm 0.04$  & $84.86 \pm 0.07$  & $\underline{85.13 \pm 0.06}$ & $\mathbf{85.30 \pm 0.09}$  \\
    Otto $\uparrow$         & $81.50 \pm 0.06$  & $63.34 \pm 0.08$  & $\underline{71.01 \pm 0.08}$ & $\mathbf{73.04 \pm 0.07}$  \\
    Cardio $\uparrow$       & $73.54 \pm 0.05$  & $72.88 \pm 0.14$  & $\underline{72.94 \pm 0.11}$ & $\mathbf{72.97 \pm 0.08}$  \\
    Insurance $\uparrow$    & $92.00 \pm 0.03$  & $92.21 \pm 0.04$  & $\underline{92.69 \pm 0.06}$ & $\mathbf{92.77 \pm 0.11}$  \\
    Forest Cov. $\uparrow$  & $96.42 \pm 0.06$  & $82.08 \pm 0.07$  & $\underline{83.84 \pm 0.08}$ & $\mathbf{85.61 \pm 0.04}$  \\
    \bottomrule
\end{tabular}
\end{table}

With respect to model usability, dynamic conditioning allows a single model to handle a variety of tasks without compromising performance. Notably, performance metrics for our multi-tasking model, \textit{\modelname{}~II} (with dynamic conditioning), align closely with those of our specific data generation model, \textit{\modelname{}~I} (without dynamic conditioning), as shown in Table~\ref{tab:ml_res}. Dynamic conditioning supports not only synthetic data generation and missing data imputation but also ``prompted data generation'' - a scenario where a synthetic subset of features is generated based on a known subset of variables. This method is particularly useful in settings where data collection is challenging, enhancing data augmentation in ML projects, improving customer profiling, or acting as a simulated environment in reinforcement learning, which accelerates data-generation efforts.

\subsubsection{Missing data imputation}

Here, we report the results of our experiments on evaluating the models' ability to impute missing values. Specifically, in this task, the generative model utilizes the available data to condition the generation of data for the missing entries. 

\begin{algorithm}
\caption{Evaluate Imputation Model Performance}
\label{alg:imputation}
\begin{algorithmic}[1]
\For{each dataset in the benchmark}
    \For{each missing value type $\left\lbrace \text{MCAR}, \text{MAR}, \text{MNAR}\right\rbrace$}
        \For{each percentage of missing values in $\left\lbrace 10\%, 25\%, 40\% \right\rbrace$}
            \State Train the generative model on $30\%$ ``imputation training'' subset.
            \State Impute the missing values in the $30\%$ ``imputation testing/discriminative training'' subset.
            \State Train the ML discriminative model (e.g., \textit{XGBoost}, \textit{CatBoost}, \textit{LightGBM} and \textit{MLP}) on the imputed data.
            \State Evaluate the ML discriminative model on the hold-out $30\%$ subset.
        \EndFor
    \EndFor
\EndFor
\end{algorithmic}
\end{algorithm}

The evaluation of imputed data quality considers three possible scenarios for missing data: Missing Completely at Random (MCAR), Missing at Random (MAR), and Missing Not at Random (MNAR). For each scenario, we first divide the dataset into three subsets: $40\%$ for ``imputation training'', $30\%$ for ``imputation testing/discriminative training''  and $30\%$ as a hold-out set. Next, we generate three versions of the imputation training and imputation test/discriminative train splits, each containing an increasing proportion of missing data ($10\%$, $25\%$, and $40\%$), while keeping the hold-out set intact. To assess the quality of the imputed data in terms of machine learning efficiency, we follow the steps outlined in Alg.~\ref{alg:imputation}. For the MAR and MNAR scenarios, the missing data are generated using the code used by \citep{jarrett2022hyperimpute} and \cite{Muzellec2020Missing}.

\begin{table}[ht!]
\caption{\textit{Missing imputation} analysis in terms of \textit{Machine Learning efficiency} when the missing information is MCAR. Classification tasks use F1-score, and regression tasks use MSE, indicated by up/down arrows for maximization/minimization of the metric. Cross-validation mean and standard deviation are shown for each dataset-model pair. Best and second-best results are highlighted in bold and underline, respectively.}
\label{tab:missing:mcar}
\resizebox{0.95\linewidth}{!}{
\begin{tabular}{l|c|ccccccc}
    \toprule
Dataset & $\%$ Missing & missForest & GAIN & HyperImpute & Miracle & TabCSDI & \modelname \\
 \midrule
                          & $10\%$ & $83.16 \pm 0.02$ & $82.44 \pm 0.02$   & $83.36 \pm 0.04$  & $\underline{83.39 \pm 0.03}$  & $83.30 \pm 0.05$  & $\mathbf{83.79 \pm 0.03}$ \\
HELOC $\uparrow$          & $25\%$ & $82.80 \pm 0.04$ & $81.92 \pm 0.04$   & $83.23 \pm 0.04$  & $82.85 \pm 0.04$  & $\underline{83.27 \pm 0.04}$  & $\mathbf{83.54 \pm 0.04}$ \\
                          & $40\%$ & $82.59 \pm 0.04$ & $81.74 \pm 0.03$   & $\underline{82.97 \pm 0.03}$  & $82.74 \pm 0.03$  & $82.67 \pm 0.03$  & $\mathbf{83.41 \pm 0.04}$ \\
\midrule
                          & $10\%$ & $84.49 \pm 0.03$ & $\underline{84.66 \pm 0.03}$   & $84.47 \pm 0.05$  & $84.60 \pm 0.03$  & $84.48 \pm 0.03$ & $\mathbf{84.76 \pm 0.03}$ \\
Churn $\uparrow$          & $25\%$ & $\underline{84.13 \pm 0.03}$ & $83.45 \pm 0.05$   & $83.89 \pm 0.05$  & $84.08 \pm 0.03$  & $83.92 \pm 0.03$  & $\mathbf{84.51 \pm 0.04}$ \\
                          & $40\%$ & $83.08 \pm 0.04$ & $83.22 \pm 0.04$   & $83.14 \pm 0.04$  & $82.52 \pm 0.04$  & $\underline{83.45 \pm 0.04}$  & $\mathbf{84.18 \pm 0.04}$ \\
\midrule
                          & $10\%$ & $0.182 \pm 0.002$ & $0.182 \pm 0.002$ & $\underline{0.173 \pm 0.002}$ & $0.174 \pm 0.002$ & $0.175 \pm 0.002$ & $\mathbf{0.170 \pm 0.002}$ \\
Cal. Hous. $\downarrow$   & $25\%$ & $0.229 \pm 0.002$ & $0.249 \pm 0.002$ & $0.190 \pm 0.002$ & $0.196 \pm 0.002$ & $\underline{0.188 \pm 0.003}$ & $\mathbf{0.183 \pm 0.002}$ \\
                          & $40\%$ & $0.269 \pm 0.004$ & $0.306 \pm 0.004$ & $\underline{0.218 \pm 0.003}$ & $0.269 \pm 0.004$ & $0.258 \pm 0.004$ & $\mathbf{0.210 \pm 0.003}$ \\
\midrule
                          & $10\%$ & $0.109 \pm 0.002$ & $0.110 \pm 0.002$ & $0.112 \pm 0.002$ & $\underline{0.106 \pm 0.002}$ & $0.110 \pm 0.002$ & $\mathbf{0.105 \pm 0.002}$ \\
House Sales $\downarrow$  & $25\%$ & $0.118 \pm 0.002$ & $0.131 \pm 0.003$ & $0.120 \pm 0.003$ & $0.127 \pm 0.002$ & $\underline{0.117 \pm 0.003}$ & $\mathbf{0.115 \pm 0.002}$ \\
                          & $40\%$ & $0.155 \pm 0.003$ & $0.145 \pm 0.003$ & $0.156 \pm 0.003$ & $0.130 \pm 0.003$ & $\underline{0.128 \pm 0.003}$ & $\mathbf{0.119 \pm 0.002}$ \\
    \bottomrule
\end{tabular}}
\end{table}

\begin{table}[ht!]
\caption{\textit{Missing imputation} analysis in terms of \textit{Machine Learning efficiency} when the missing information is MAR. Classification tasks use F1-score, and regression tasks use MSE, indicated by up/down arrows for maximization/minimization of the metric. Cross-validation mean and standard deviation are shown for each dataset-model pair. Best and second-best results are highlighted in bold and underline, respectively.}
\label{tab:missing:mar}
\resizebox{0.95\linewidth}{!}{
\begin{tabular}{l|c|ccccccc}
    \toprule
Dataset & $\%$ Missing & missForest & GAIN & HyperImpute & Miracle & TabCSDI & \modelname \\
 \midrule
                          & $10\%$ & $83.13 \pm 0.02$ & $82.39 \pm 0.04$ & \underline{$83.34 \pm 0.02$} & $83.30 \pm 0.03$ & $83.25 \pm 0.03$ & $\mathbf{83.78 \pm 0.02}$\\
HELOC $\uparrow$          & $25\%$ & $82.75 \pm 0.03$ & $81.88 \pm 0.02$ & $83.15 \pm 0.03$ & $82.81 \pm 0.02$ & \underline{$83.18 \pm 0.01$} & $\mathbf{83.53 \pm 0.04}$\\
                          & $40\%$ & $82.42 \pm 0.02$ & $81.64 \pm 0.04$ & \underline{$82.86 \pm 0.04$} & $82.52 \pm 0.02$ & $82.59 \pm 0.05$ & $\mathbf{83.40 \pm 0.04}$\\
\midrule 
                          & $10\%$ & $84.46 \pm 0.04$ & $84.55 \pm 0.03$ & $84.42 \pm 0.02$ & \underline{$84.58 \pm 0.04$} & $84.44 \pm 0.03$ &  $\mathbf{84.76 \pm 0.03}$\\
Churn $\uparrow$          & $25\%$ & \underline{$84.05 \pm 0.03$} & $83.39 \pm 0.02$ & $83.79 \pm 0.02$ & $84.00 \pm 0.02$ & $83.58 \pm 0.02$ &  $\mathbf{84.50 \pm 0.02}$\\
                          & $40\%$ & $82.91 \pm 0.05$ & $83.11 \pm 0.01$ & $83.00 \pm 0.03$ & $82.40 \pm 0.04$ & \underline{$83.31 \pm 0.03$} &  $\mathbf{84.17 \pm 0.04}$\\
\midrule 
                          & $10\%$ & $0.182 \pm 0.003$ & $0.183 \pm 0.004$ & \underline{$0.173 \pm 0.003$} & $0.174 \pm 0.003$ & $0.175 \pm 0.004$ & $\mathbf{0.171 \pm 0.003}$ \\
Cal. Hous. $\downarrow$   & $25\%$ & $0.230 \pm 0.002$ & $0.251 \pm 0.004$ & $0.191 \pm 0.002$ & $0.197 \pm 0.004$ & \underline{$0.190 \pm 0.005$} & $\mathbf{0.183 \pm 0.002}$ \\
                          & $40\%$ & $0.272 \pm 0.002$ & $0.308 \pm 0.005$ & \underline{$0.222 \pm 0.003$} & $0.274 \pm 0.002$ & $0.264 \pm 0.004$ & $\mathbf{0.211 \pm 0.003}$ \\
\midrule 
                          & $10\%$ & $0.110 \pm 0.004$ & $0.112 \pm 0.002$ & $0.114 \pm 0.004$ & \underline{$0.106 \pm 0.002$} & $0.111 \pm 0.004$ & $\mathbf{0.105 \pm 0.002}$ \\
House Sales $\downarrow$  & $25\%$ & \underline{$0.120 \pm 0.003$} & $0.134 \pm 0.002$ & $0.122 \pm 0.004$ & $0.130 \pm 0.002$ & $0.121 \pm 0.003$ & $\mathbf{0.116 \pm 0.003}$ \\
                          & $40\%$ & $0.158 \pm 0.003$ & $0.149 \pm 0.001$ & $0.160 \pm 0.003$ & $0.134 \pm 0.003$ & \underline{$0.132 \pm 0.003$} & $\mathbf{0.120 \pm 0.002}$ \\
    \bottomrule
\end{tabular}}
\end{table}

\begin{table}[ht!]
\caption{\textit{Missing imputation} analysis in terms of \textit{Machine Learning efficiency} when the missing information is MNAR. Classification tasks use F1-score, and regression tasks use MSE, indicated by up/down arrows for maximization/minimization of the metric. Cross-validation mean and standard deviation are shown for each dataset-model pair. Best and second-best results are highlighted in bold and underline, respectively.}
\label{tab:missing:mnar}
\resizebox{0.95\linewidth}{!}{
\begin{tabular}{l|c|ccccccc}
    \toprule
Dataset & $\%$ Missing & missForest & GAIN & HyperImpute & Miracle & TabCSDI & \modelname \\
 \midrule
                          & $10\%$ & $83.09 \pm 0.03$ & $82.35 \pm 0.03$ & \underline{$83.31 \pm 0.04$} & $83.25 \pm 0.04$ & $83.23 \pm 0.04$ & $\mathbf{83.75 \pm 0.04}$ \\
HELOC $\uparrow$          & $25\%$ & $82.59 \pm 0.03$ & $81.78 \pm 0.02$ & $83.04 \pm 0.05$ & $82.69 \pm 0.05$ & \underline{$83.07 \pm 0.05$} & $\mathbf{83.47 \pm 0.03}$ \\
                          & $40\%$ & $82.12 \pm 0.04$ & $81.43 \pm 0.03$ & \underline{$82.67 \pm 0.03$} & $82.35 \pm 0.03$ & $82.31 \pm 0.04$ & $\mathbf{83.31 \pm 0.04}$ \\
\midrule 
                          & $10\%$ & $84.42 \pm 0.04$ & \underline{$84.51 \pm 0.05$} & $84.36 \pm 0.04$ & $84.49 \pm 0.03$ & $84.38 \pm 0.04$ &  $\mathbf{84.72 \pm 0.04}$ \\
Churn $\uparrow$          & $25\%$ & \underline{$83.95 \pm 0.04$} & $83.27 \pm 0.04$ & $83.66 \pm 0.05$ & $83.86 \pm 0.04$ & $83.72 \pm 0.03$ &  $\mathbf{84.42 \pm 0.03}$ \\
                          & $40\%$ & $82.72 \pm 0.03$ & $82.93 \pm 0.05$ & $82.80 \pm 0.05$ & $82.23 \pm 0.03$ & \underline{$83.09 \pm 0.04$} &  $\mathbf{84.04 \pm 0.05}$ \\
\midrule 
                          & $10\%$ & $0.184 \pm 0.003$ & $0.185 \pm 0.004$ & $0.177 \pm 0.004$ & $0.178 \pm 0.003$ & \underline{$0.176 \pm 0.003$} &  $\mathbf{0.173 \pm 0.003}$\\
Cal. Hous. $\downarrow$   & $25\%$ & $0.234 \pm 0.004$ & $0.255 \pm 0.003$ & $0.200 \pm 0.003$ & $0.205 \pm 0.002$ & \underline{$0.194 \pm 0.004$} & $\mathbf{0.187 \pm 0.003}$ \\
                          & $40\%$ & $0.285 \pm 0.003$ & $0.318 \pm 0.002$ & \underline{$0.239 \pm 0.004$} & $0.289 \pm 0.003$ & $0.279 \pm 0.002$ & $\mathbf{0.220 \pm 0.002}$ \\
\midrule 
                          & $10\%$ & \underline{$0.111 \pm 0.002$} & $0.116 \pm 0.004$ & $0.117 \pm 0.004$ & $0.112 \pm 0.002$ & $0.115 \pm 0.003$ & $\mathbf{0.108 \pm 0.003}$\\
House Sales $\downarrow$  & $25\%$ & $0.127 \pm 0.001$ & $0.140 \pm 0.002$ & $0.129 \pm 0.003$ & $0.137 \pm 0.004$ & \underline{$0.126 \pm 0.002$} & $\mathbf{0.120 \pm 0.002}$\\
                          & $40\%$ & $0.169 \pm 0.002$ & $0.160 \pm 0.003$ & $0.171 \pm 0.003$ & $0.143 \pm 0.003$ & \underline{$0.140 \pm 0.004$} & $\mathbf{0.126 \pm 0.003}$\\
    \bottomrule
\end{tabular}}
\end{table}

The results of the experiments for the MCAR, MAR, and MNAR scenarios are presented in Tables \ref{tab:missing:mcar}, \ref{tab:missing:mar}, \ref{tab:missing:mnar}, respectively. As the assumption about missingness becomes more complex (MCAR $\rightarrow$ MAR $\rightarrow$ MNAR), all the models included in the comparison exhibit a very slight degradation in performance. However, the findings remain consistent across all missing data scenarios.

Our experiments indicate that \textit{\modelname{}} tends to outperform the baseline models considered in this evaluation: \textit{missForest} \citep{missForest}, \textit{GAIN}  \citep{yoon2018gain}, \textit{HyperImpute} \citep{jarrett2022hyperimpute}, \textit{Miracle} \citep{kyono2021miracle}, and \textit{TabCSDI} \citep{zheng2022diffusion}. Across all levels of missing data, \textit{\modelname{}} generally achieves better performance on average, with its advantage becoming more pronounced as the proportion of missing data increases.

\subsection{Statistical Similarity} 

In this task, we compare synthetic and real data accounts based on individual and joint feature distributions. We report only the top three models with the best ML efficiency. This approach is motivated by our findings, which show a strong association between the ML efficiency of synthetic data and their ability to replicate the statistical properties of real data. In other words, synthetic data with higher ML utility tends to better reproduce both individual and joint feature distributions.

\begin{table}[H]
\caption{Statistical similarity details}
    \label{tab:stat_distr_details}
    \resizebox{0.95\textwidth}{!}{
    \begin{tabular}{ccc}
     (a) Average Wasserstein Distance &  (b) Average Jensen-Shannon Distance &  (c) Average L2 Dist. Correlation Matrix \\
     
    \begin{tabular}{cccc}
    \toprule
    Dataset	& Tabsyn & TabDDPM & \modelname{} \\    
    \midrule
    HELOC	    & $\underline{0.25 \pm 0.03}$   & $0.29 \pm 0.02$   & $\mathbf{0,15 \pm 0.02}$ \\
    Churn	    & $\mathbf{0.15 \pm 0.02}$	& $0.23 \pm 0.02$	& $\underline{0.17 \pm 0.02}$ \\
    Gas	        & $\mathbf{0.21 \pm 0.04}$	& $0.65 \pm 0.06$	& $\underline{0.23 \pm 0.02}$ \\
    Cal.Hous	& $\underline{0.31 \pm 0.03}$	& $0.34 \pm 0.04$	& $\mathbf{0.22 \pm 0.02}$ \\
    House Sales	& $0.32 \pm 0.02$	& $\underline{0.28 \pm 0.03}$	& $\mathbf{0.15 \pm 0.01}$ \\
    Adult Inc.  & $\mathbf{0.18 \pm 0.03}$	& $0.24 \pm 0.02$	& $\underline{0.19 \pm 0.02}$ \\
    Otto	    & $\underline{0.25 \pm 0.02}$	& $0.27 \pm 0.03$	& $\mathbf{0.16 \pm 0.02}$ \\
    Cardio	    & $\mathbf{0.23 \pm 0.04}$	& $0.26 \pm 0.03$	& $\underline{0.24 \pm 0.03}$ \\
    Insurance	& $\mathbf{0.19 \pm 0.02}$	& $0.22 \pm 0.02$	& $\underline{0.20 \pm 0.03}$ \\
    Forest Cov. & $0.41 \pm 0.05$	& $\underline{0.35 \pm 0.04}$	& $\mathbf{0.18 \pm 0.02}$  \\
    \midrule
    Av. Rank    & $\underline{1.7}$ & $2.8$ & $\mathbf{1.5}$ \\
    \bottomrule
    \end{tabular} & 
    \begin{tabular}{cccc}
    \toprule
    Dataset	& Tabsyn & TabDDPM & \modelname{} \\    
    \midrule
    HELOC	    & $\underline{0.15 \pm 0.02}$   & $0.20 \pm 0.02$   & $\mathbf{0,08 \pm 0.01}$ \\
    Churn	    & $\mathbf{0.07 \pm 0.01}$	& $0.15 \pm 0.02$	& $\underline{0.09 \pm 0.01}$ \\
    Gas	        & NA	& NA	& NA \\
    Cal.Hous	& NA	& NA	& NA \\
    House Sales	& $\underline{0.22 \pm 0.02}$	& $0.26 \pm 0.03$	& $\mathbf{0.09 \pm 0.01}$ \\
    Adult Inc.  & $\mathbf{0.07 \pm 0.01}$	& $0.14 \pm 0.02$	& $\underline{0.08 \pm 0.02}$ \\
    Otto	    & NA	& NA 	& NA \\
    Cardio	    & $\mathbf{0.10 \pm 0.01}$	& $0.14 \pm 0.02$	& $\underline{0.11 \pm 0.01}$ \\
    Insurance	& $\mathbf{0.09 \pm 0.01}$	& $0.13 \pm 0.01$	& $\underline{0.10 \pm 0.01}$ \\
    Forest Cov. & $0.24 \pm 0.03$	& $\underline{0.16 \pm 0.02}$	& $\mathbf{0.08 \pm 0.01}$  \\
    \midrule
    Av. Rank    & $\mathbf{1.6}$ & $2.9$ & $\mathbf{1.6}$ \\
    \bottomrule
    \end{tabular} &
    \begin{tabular}{cccc}
    \toprule
    Dataset	& Tabsyn & TabDDPM & \modelname{} \\    
    \midrule
    HELOC	    & $\underline{0.036 \pm 0.004}$   & $0.040 \pm 0.004$   & $\mathbf{0,025 \pm 0.003}$ \\
    Churn	    & $\mathbf{0.027 \pm 0.003}$	& $0.034 \pm 0.003$	& $\underline{0.028 \pm 0.003}$ \\
    Gas	        & $\mathbf{0.032 \pm 0.003}$	& $0.078 \pm 0.007$	& $\underline{0.034 \pm 0.004}$ \\
    Cal.Hous	& $\underline{0.042 \pm 0.004}$	& $0.045 \pm 0.005 $	& $\mathbf{0.031 \pm 0.003}$ \\
    House Sales	& $0.044 \pm 0.004$	& $\underline{0.039 \pm 0.004}$	& $\mathbf{0.024 \pm 0.003}$ \\
    Adult Inc.  & $\mathbf{0.029 \pm 0.003}$	& $0.036 \pm 0.004$	& $\underline{0.030 \pm 0.003}$ \\
    Otto	    & $\underline{0.037\pm 0.004}$	& $0.041\pm 0.004$ 	& $\mathbf{0.026\pm 0.003}$ \\
    Cardio	    & $\mathbf{0.032 \pm 0.003}$	& $0.035 \pm 0.003$	& $\underline{0.033 \pm 0.003}$ \\
    Insurance	& $\underline{0.031 \pm 0.003}$	& $0.033 \pm 0.003$	& $\mathbf{0.029 \pm 0.003}$ \\
    Forest Cov. & $0.052 \pm 0.004$	& $\underline{0.046 \pm 0.004}$	& $\mathbf{0.031 \pm 0.003}$  \\
    \midrule
    Av. Rank    & $\underline{1.8}$ & $2.8$ & $\mathbf{1.4}$ \\
    \bottomrule
    \end{tabular} \\
    \end{tabular}}
\end{table}

\begin{figure}[H]
    \centering
    \includegraphics[width=0.87\linewidth]{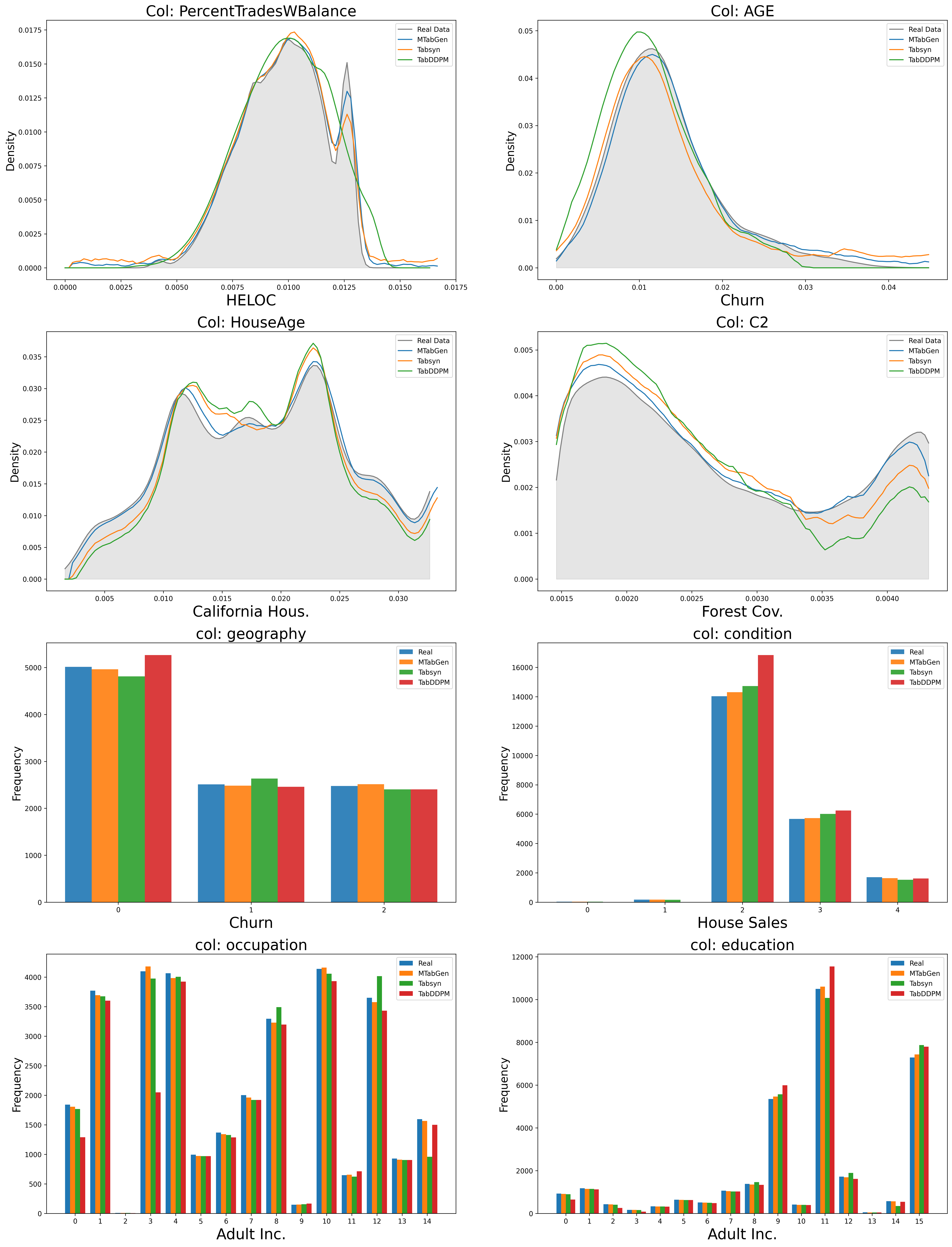}
    \caption{Comparison between distribution of real and synthetic data. The two top rows present four examples of numerical columns from various datasets in our benchmark, while the two bottom rows contain examples of categorical features.}
    \label{fig:stats_plots}
\end{figure}

Table~\ref{tab:stat_distr_details}-a shows the average Wasserstein Distance between synthetic and real numerical data distributions. Specifically, the Wasserstein distance is calculated for each numerical column between the real data and the synthetic data generated by \textit{Tabsys}, \textit{TabDDPM}, and \textit{\modelname{}}. For each generative model, the final dataset results are the average of these distances across all numerical columns in the dataset. \textit{\modelname{}} consistently performs better than \textit{Tabsys} and \textit{TabDDPM}. This advantage is more pronounced in datasets where there is a larger difference in ML efficiency, such as \textit{HELOC}, \textit{California Housing}, \textit{House Sales} or \textit{Forest Cover Type} dataset. In contrast, in datasets like \textit{Insurance}, where all models have similar ML efficiency, the Wasserstein distances are also comparable. A qualitative analysis of the results for some of the columns can be found in the four plots in the first two rows plots of Fig.~\ref{fig:stats_plots}. These plots compare the distributions of real data with those of the synthetic data generated by the different models.

Similarly, Table~\ref{tab:stat_distr_details}-b shows the average Jensen-Shannon distance between synthetic and real categorical data distributions. \textit{\modelname{}} and \textit{Tabsyn} outperform \textit{TabDDPM}, consistent with the findings of \citep{zhang2024mixedtype}, where \textit{Tabsyn} achieved better results than \textit{TabDDPM} on categorical data. Also the Jensen-Shannon distance appears to be strongly related to ML efficiency. For example, in datasets like \textit{HELOC}, \textit{House Sales} or \textit{Forest Cover Type}, where ML efficiency of \textit{\modelname{}} is significantly better than the other baselines, it also has a lower Jensen-Shannon distance. In the \textit{Insurance} dataset, where all baselines have similar ML efficiency, the Jensen-Shannon distances are also similar. Notably, while \textit{\modelname{}} and \textit{Tabsyn} show similar overall results, a qualitative analysis of per-column results reveals that \textit{\modelname{}} tends to excel when the number of classes in the categorical variable increases, as shown in the last two rows plots of Fig.~\ref{fig:stats_plots}.

\begin{figure}[ht!]
    \centering
    \includegraphics[width=0.9\textwidth]{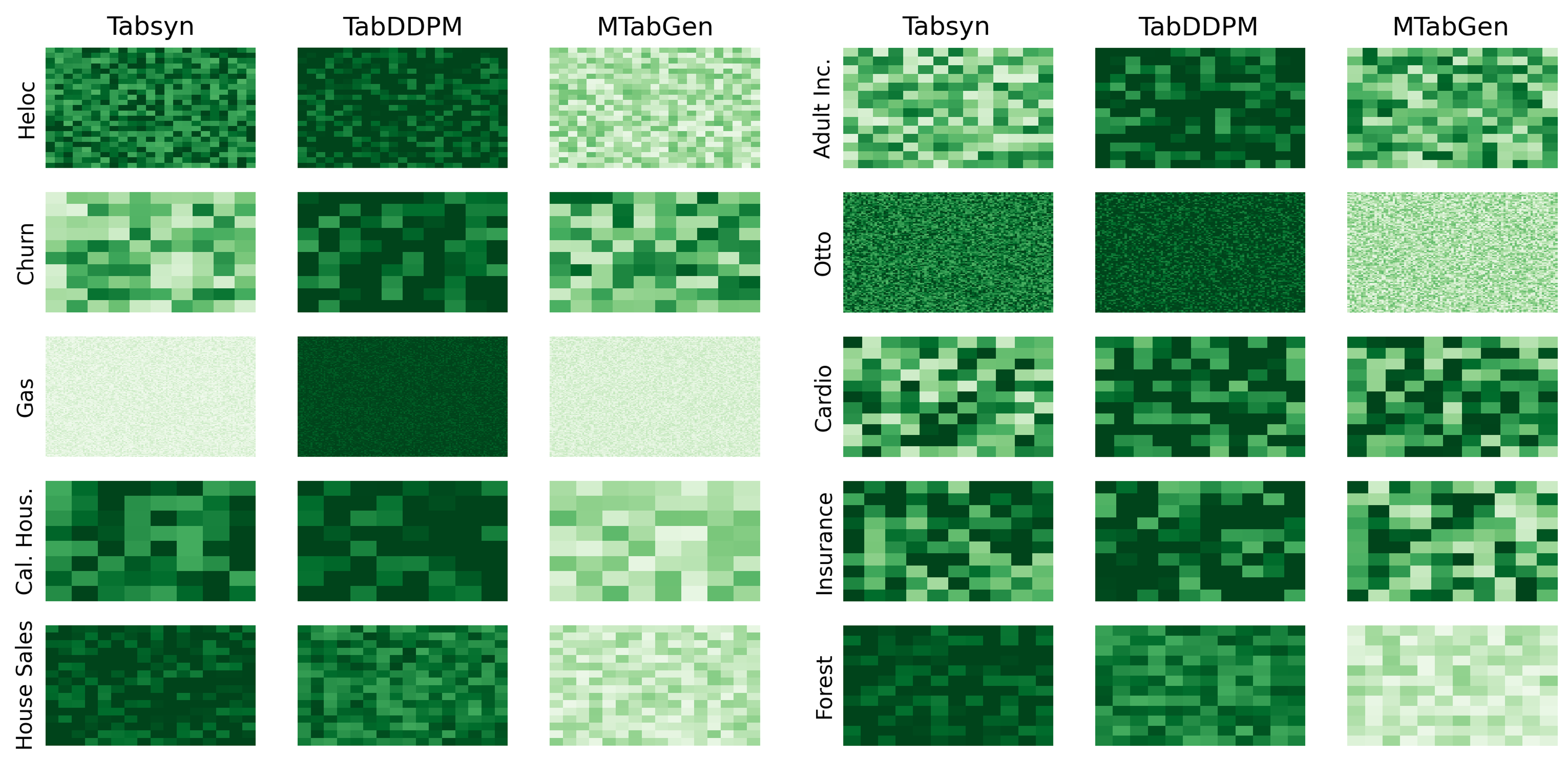}
    \caption{L2 distance between correlation matrices computed on real and synthetic data. More intense green color means higher difference between the real and synthetic correlation values.} 
    \label{fig:l2_corr}
\end{figure}

Table~\ref{tab:stat_distr_details}-c shows the average L2 distance between the two correlation matrices computed on real and synthetic data. Fig.~\ref{fig:l2_corr}, instead, shows the L2 distance details across all the datasets in the benchmark. In Fig.~\ref{fig:l2_corr} more intense green color means higher difference between the real and synthetic correlation values. Even more than in the previous statistical measures, there exist an association between ML efficiency and L2 distance between correlation matrices. In datasets like \textit{HELOC}, \textit{California Housing}, \textit{House Sales} or \textit{Forest Cover Type} where the ML efficiency of \textit{\modelname{}} is notably better than the one of \textit{Tabsyn} and \textit{TabDDPM} the corresponding heatmap of   \textit{\modelname{}} are more lighter, e.g. they shown a smaller error in the correlation estimation. Nevertheless in datasets like \textit{Cardio} or \textit{Insurance} where all the models perform similarly,  the heatmaps do not show resignable differences.

\subsection{Privacy risk}

In this section, we delve deeper into the privacy risk associated with synthetic data. As previously mentioned, DCR is defined by the Euclidean distance between any synthetic record and its closest real neighbor. Ideally, a higher DCR indicates a lower privacy risk. However, out-of-distribution data (random noise) can also result in high DCR. Therefore, to maintain ecological validity, DCR should be evaluated alongside the ML efficiency metric. For this reason, our privacy risk evaluation includes only \textit{Tabsyn}, \textit{TabDDPM}, and \textit{\modelname{}}. These models have superior ML efficiency and are more likely to pose a privacy risk (lower DCR) because they closely mimic the original data.

\begin{table}[ht!]
\caption{Comparison of privacy risk ($\uparrow$) and ML efficiency ($\uparrow$ for F1-score, $\downarrow$ for MSE) for \textit{Tabsyn}, \textit{TabDDPM}, and \textit{\modelname{}} for each dataset. Privacy risk is evaluated using the Distance to Closest Record; higher values indicate a lesser risk of privacy breach. The results are divided based on a 5\% ML efficiency threshold relative to the performance improvements of \textit{\modelname{}} compared to \textit{Tabsyn} and \textit{TabDDPM}. In cases where \textit{\modelname{}} achieves comparable ML efficiency to the baselines, the privacy risk is similar. However, in scenarios where \textit{\modelname{}} outperforms the other baselines, there is higher privacy risk (i.e lower value of DCR distance), as expected, since better ML efficiency is related with closer statistical fidelity of the generated data. }
\label{tab:risk}
\centering

\begin{minipage}[t]{.44\linewidth}
\centering
\caption*{Tabsyn and \modelname{}}
\begin{tabular}{l|cc|cc}
    \toprule
    \multirow{2}{*}{Dataset} & \multicolumn{2}{c|}{Tabsyn} & \multicolumn{2}{c}{\modelname{}} \\
    \cmidrule{2-3} \cmidrule{4-5}
    & Risk $\uparrow$ & ML eff. & Risk $\uparrow$ & ML eff. \\
    \midrule
    HELOC        & $\mathbf{0.35}$ & $79.24$ $\uparrow$ & $0.25$ & $\mathbf{82.91} $ $\uparrow$ \\
    Churn        & $0.06$ & $\mathbf{84.60}$ $\uparrow$ & $\mathbf{0.09}$ & $84.43$ $\uparrow$ \\
    Gas          & $\mathbf{0.21}$ & $98.66$ $\uparrow$ & $0.19$ & $\mathbf{98.80}$ $\uparrow$ \\
    Adult Inc.   & $\mathbf{0.13} $ & $84.70$ $\uparrow$ & $0.11$ & $\mathbf{85.30}$ $\uparrow$ \\
    Cardio       & $0.35$ & $72.90$ $\uparrow$ & $\mathbf{0.41}$ & $\mathbf{72.97}$ $\uparrow$ \\
    Insurance    & $\mathbf{0.12}$ & $92.20$ $\uparrow$ & $\mathbf{0.12}$ & $\mathbf{92.77}$ $\uparrow$ \\
    \midrule
    \multicolumn{5}{c}{
    5\% Relative ML eff. 
     Threshold
    } 
    \\
    \midrule
    Cal. Hous.   & $\mathbf{0.81}$ & $0.256$ $\downarrow$ & $0.18$ & $\mathbf{0.224}$ $\downarrow$ \\
    House Sales  & $\mathbf{0.16}$ & $0.148$ $\downarrow$ & $0.10$ & $\mathbf{0.121}$ $\downarrow$ \\
    Otto         & $\mathbf{1.08}$ & $67.14$ $\uparrow$ & $0.12$ & $\mathbf{73.04}$ $\uparrow$ \\
    Forest Cov.  & $\mathbf{0.85}$ & $74.83$ $\uparrow$ & $0.35$ & $\mathbf{85.61}$ $\uparrow$ \\
    \bottomrule
\end{tabular}
\end{minipage} \hspace{1cm}
\begin{minipage}[t]{.44\linewidth}
\centering
\caption*{TabDDPM and \modelname{}}
\begin{tabular}{l|cc|cc}
    \toprule
    \multirow{2}{*}{Dataset} & \multicolumn{2}{c|}{TabDDPM} & \multicolumn{2}{c}{\modelname{}} \\
    \cmidrule{2-3} \cmidrule{4-5}
    & Risk $\uparrow$ & ML eff. & Risk $\uparrow$ & ML eff. \\
    \midrule
    Churn        & $\mathbf{0.09}$ & $83.62$ $\uparrow$ & $\mathbf{0.09}$ & $\mathbf{84.43}$ $\uparrow$ \\
    Adult Inc.   & $\mathbf{0.15}$ & $84.86$ $\uparrow$ & $0.11$ & $\mathbf{85.30}$ $\uparrow$ \\
    Cardio       & $\mathbf{0.41}$ & $72.88$ $\uparrow$ & $\mathbf{0.41}$ & $\mathbf{72.97}$ $\uparrow$ \\
    Insurance    & $\mathbf{0.12}$ & $92.21$ $\uparrow$ & $\mathbf{0.12}$ & $\mathbf{92.77}$ $\uparrow$ \\
    Forest Cov.  & $\mathbf{0.47}$ & $82.08$ $\uparrow$ & $0.35$ & $\mathbf{85.61}$ $\uparrow$ \\
    \midrule
    \multicolumn{5}{c}{
    5\% Relative ML eff. 
     Threshold} \\
    \midrule
    HELOC        & $\mathbf{2.75}$ & $76.69$ $\uparrow$ & $0.25$ & $\mathbf{82.91}$ $\uparrow$ \\
    Gas          & $\mathbf{3.15}$ & $65.51$ $\uparrow$ & $0.19$ & $\mathbf{98.80}$ $\uparrow$ \\
    Cal. Hous.   & $\mathbf{1.28}$ & $0.272$ $\downarrow$ & $0.18$ & $\mathbf{0.224}$ $\downarrow$ \\
    House Sales  & $\mathbf{0.34}$ & $0.145$ $\downarrow$ & $0.10$ & $\mathbf{0.121}$ $\downarrow$ \\
    Otto         & $\mathbf{2.57}$ & $63.34$ $\uparrow$ & $0.12$ & $\mathbf{73.04}$ $\uparrow$ \\
    \bottomrule
\end{tabular}
\end{minipage}

\end{table}

Table~\ref{tab:risk} presents our evaluation results regarding the privacy risk metric, emphasizing the trade-off between ML efficiency and privacy guarantees. Based on a 5\% threshold for relative improvement in ML efficiency of \textit{\modelname{}} over \textit{Tabsyn} and \textit{TabDDPM}, the results are categorized to illustrate two key points: (1) Improvements in ML efficiency correlate with increased privacy risks. (2) When \textit{\modelname{}}'s performance improvement over the baselines is less than 5\%, no significant increase in privacy risk is observed. However, when \textit{\modelname{}}'s performance is substantially better, the comparison becomes irrelevant because the synthetic data generated by the baselines deviate significantly from real data in terms of ML utility and statistical properties, resulting in a high DCR due to poor alignment with real data properties.

\begin{figure}[ht!]
    \centering
    \includegraphics[width=0.95\linewidth]{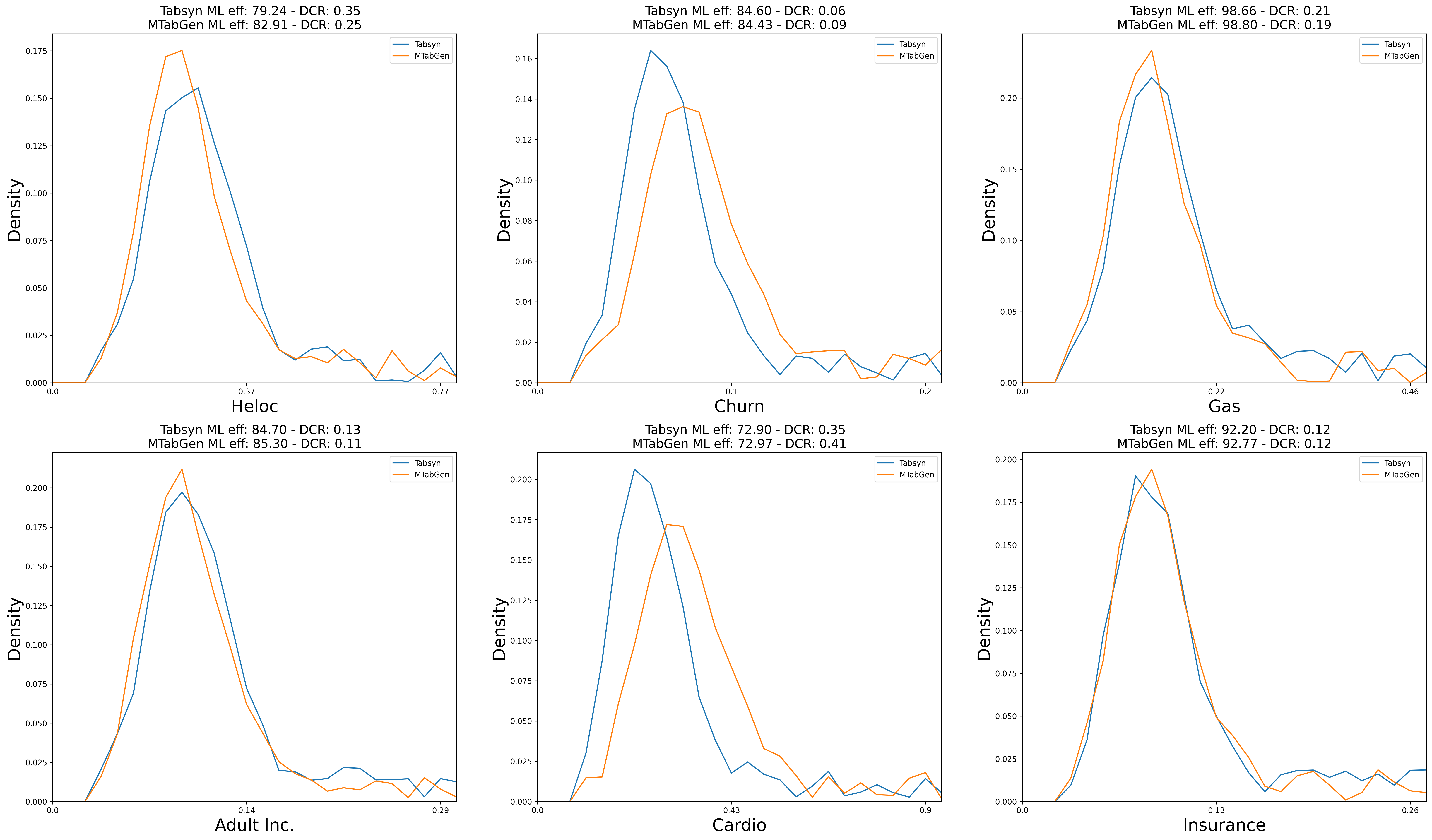}
    \caption{Distance to Closest Record (DCR) for \textit{Tabsyn} and \textit{\modelname{}}.} 
    \label{fig:dcr_tabsyn_mtabgen}
\end{figure}

\begin{figure}[ht!]
    \centering
    \includegraphics[width=0.95\linewidth]{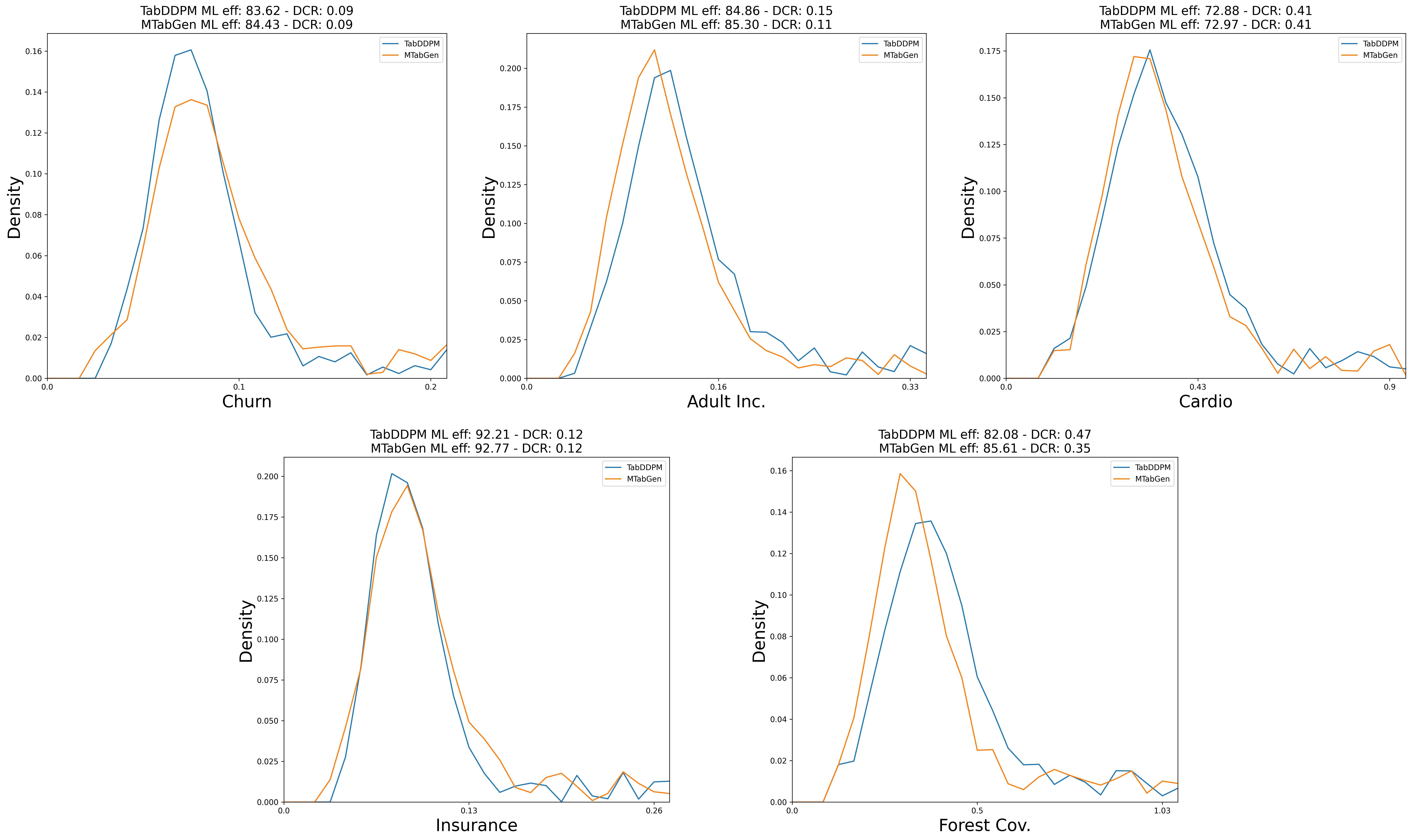}
    \caption{Distance to Closest Record (DCR) for \textit{TabDDPM} and \textit{\modelname{}}.} 
    \label{fig:dcr_tabddpm_mtabgen}
\end{figure}

Figures~\ref{fig:dcr_tabsyn_mtabgen} and \ref{fig:dcr_tabddpm_mtabgen}  provide a qualitative comparison of the DCR distribution for \textit{\modelname{}} versus \textit{Tabsyn} and \textit{TabDDPM} when the relative improvement in ML efficiency is less than 5\%. Specifically, for each synthetic data point, we compute the distance to its closest record in the training dataset. The plots then depict the distribution of these distances. This comparison confirms that there are no significant changes or patterns, indicating that the improvement in ML efficiency does not lead to a noticeable increase in privacy risk.

To mitigate the risk of increasing privacy concerns, our framework is equipped to integrate additional privacy-preserving measures, such as differential privacy \citep{jalko2021privacy}, allowing for a better-controlled balance between data efficiency and privacy. Additionally, we conduct a sanity check to ensure that no synthetic sample perfectly matches any original sample (i.e., DCR is always greater than 0), safeguarding against direct data leakage.

\section{Conclusion}

In this paper, we introduced \textit{\modelname{}}, a diffusion model enhanced with a conditioning attention mechanism, a transformer-based encoder-decoder architecture, and dynamic masking. \textit{\modelname{}} is specifically designed for applications involving mixed-type tabular data. The transformer encoder-decoder acts as the denoising network, enabling the conditioning attention mechanism while effectively capturing and representing complex interactions and dependencies within the data. The dynamic masking feature allows \textit{\modelname{}} to handle both synthetic data generation and missing data imputation tasks within a unified framework efficiently. We proposed to train the diffusion model to regenerate masked data, enabling applications ranging from data imputation to unconditioned or conditioned synthetic data generation. This versatility makes \textit{\modelname{}} suitable for generating synthetic data to overcome privacy regulations, augment existing datasets, or mitigate class imbalances.
We evaluated \textit{\modelname{}} against established baselines across several public datasets with a diverse range of features. Our model demonstrated better overall performance in terms of ML efficiency and statistical accuracy, while maintaining privacy risks comparable to those of the baselines, particularly showing increased performance in datasets with a large number of features.

\section*{Impact Statements}

Tabular data is one of the most common structures with which to represent information (e.g. finance, health, etc.). With the present model, the ability to generate or complete records in these structures is given. This fact requires ethical and privacy considerations. The authors encourage that before sharing any type of data, whether original or generated with the proposed model, to verify that reverse-identification is impossible or prevented by regulatory means. 
In addition to these considerations, our model and methods described in the paper can also be utilized to rebalance datasets for minority groups by synthetically generating new samples conditioned on the minority class, thus aiding in fairer data representation.
Apart from this, we see no other ethical issues related to this work.

    \bibliographystyle{ACM-Reference-Format}
    \bibliography{references}

\appendix

\section{Hyperparameter tuning} \label{sec:hyperparameter}

\begin{table}[H]
    \caption{Discriminative models: hyperparameters search space.}
    \label{tab:optuna_search_space}
    \begin{tabular}{ccc}
        \toprule
        Model & Hyperparameter & Possible Values \\
        \midrule
        \multirow{10}{*}{XGBoost} & $\mathrm{max\,depth}$ & $\left[1,9\right]$ \\
        & $\mathrm{learning\,rate}$ & $\left[0.01, 1.0\right]$ \\
        & $\mathrm{estimators}$ & $\left[50, 500\right]$ \\
        & $\mathrm{min\,child\,weight}$ & $\left[1, 10\right]$\\
        & $\mathrm{gamma}$ & $\left[10^{-8}, 1\right]$  \\
        & $\mathrm{subsample}$ & $\left[0.01, 1\right]$ \\
        & $\mathrm{colsample\,bytree}$ & $\left[0.01, 1\right]$  \\
        & $\mathrm{reg\, alpha}$ & $\left[10^{-8}, 1\right]$ \\
        & $\mathrm{subsample}$ & $\left[0.01, 1\right]$ \\
        & $\mathrm{reg\,lambda}$ & $\left[10^{-8}, 1\right]$ \\
        \hline
        
        \multirow{5}{*}{CatBoost} & $\mathrm{learning\,rate}$ & LogUniform $\left[0.1, 1\right]$ \\
        & $\mathrm{l2\,leaf\,reg}$ &  LogUniform$\left[1, 100\right]$ \\
        & $\mathrm{bagging\,temperature}$ & LogUniform$\left[0.1, 20\right]$ \\
        & $\mathrm{random\,strength}$ & $\left[1.0, 2.0\right]$ \\
        & $\mathrm{depth}$ & $\left[1, 10\right]$ \\
        & $\mathrm{min\,data\,in\,leaf}$ & $\left[1, 300\right]$ \\
        \hline
        
        \multirow{11}{*}{LightGBM} & $\mathrm{max\,depth}$ & $\left[3,12\right]$ \\
        & $\mathrm{learning\,rate}$ & $\left[0.01, 1.0\right]$ \\
        & $\mathrm{estimators}$ & $\left[50, 500\right]$ \\
        & $\mathrm{num\,leaves}$ & $\left[20, 3000\right]$ \\
        & $\mathrm{min\,data\,in\,leaf}$ & $\left[200, 10000\right]$\\
        & $\mathrm{max\,bin}$ & $\left[200, 300\right]$  \\
        & $\mathrm{lambda\,l1}$ & $\left[0, 100\right]$ \\
        & $\mathrm{lambda\,l2}$ & $\left[0, 100\right]$  \\
        & $\mathrm{min\,gain\,to\,split}$ & $\left[0, 15\right]$ \\
        & $\mathrm{bagging\,fraction}$ & $\left[0.2, 0.95\right]$ \\
        & $\mathrm{feature\,fraction}$ & $\left[0.2, 0.95\right]$ \\
        \hline
        
        \multirow{5}{*}{MLP} & $\mathrm{hidden\,layers}$ & $\left[2, 4, 6, 8\right]$ \\
        & $\mathrm{latent\,space\,size}$ & $\left[64, 128, 256, 512\right]$ \\
        & $\mathrm{batch\,size}$ & $\left[64, 128, 256, 512, 1024\right]$ \\
        & $\mathrm{learning\,rate}$ & LogUniform[0.00001, 0.003] \\
        & $\mathrm{epochs}$ & $500$ \\
        \bottomrule
        
    \end{tabular}
\end{table}

\begin{table}[H]
    \caption{Generative models: hyperparameters search space.}
    \label{tab:model_search_space}
    \resizebox*{!}{0.95\textheight}{
    \begin{tabular}{ccc}
        \toprule
        Model & Hyperparameter & Possible Values \\
        \midrule
        \multirow{5}{*}{TVAE} & compress dims & $\left[32, 64, 128, 256, 512\right]$ \\
        & decompress dims & $\left[32, 64, 128, 256, 512\right]$ \\
        & embedding dim & $\left[32, 64, 128, 256, 512\right]$ \\
        & batch size & $\left[64, 128, 256, 512, 1024\right]$ \\
        & learning rate & LogUniform[0.00001, 0.003] \\
        & epochs & $500$ \\
        \hline
        
        \multirow{5}{*}{CTGAN} & generator dim & $\left[32, 64, 128, 256, 512\right]$ \\
        & discriminator dim & $\left[32, 64, 128, 256, 512\right]$ \\
        & embedding dim & $\left[32, 64, 128, 256, 512\right]$ \\
        & batch size & $\left[64, 128, 256, 512, 1024\right]$ \\
        & learning rate & LogUniform[0.00001, 0.003] \\
        & epochs & $500$ \\
        \hline

        \multirow{9}{*}{CoDi} & timesteps & $\left[ 50 \right]$ \\
        & learning rate & $\left[ 2e-03, 2e-05 \right]$ \\
        & dim(Emb(t)) & $\left[ 16, 32, 64, 128 \right]$ \\
        &  $\left\lbrace dim(h1), \right.$ & $\left\lbrace \right. \left\lbrace 16, 32, 64 \right\rbrace, $ \\
        &  $dim(h2), $ & $\left\lbrace 32, 64, 128 \right\rbrace, $ \\
        &  $\left. dim(h3) \right\rbrace$ & $\left\lbrace 128, 256, 512 \right\rbrace \left.\right\rbrace$ \\
        & $\lambda_C$ & $\left[ 0.2, 0.3, \dots, 0.8 \right]$ \\
        & $\lambda_D$ & $\left[ 0.2, 0.3, \dots, 0.8 \right]$ \\
        & epochs & $500$ \\
        \hline

        \multirow{9}{*}{Tabsyn} & VAE-n heads & $\left[1, 2, 4\right]$ \\
        & VAE-Factor & $\left[16, 32, 64, 128\right]$ \\
        & VAE-Layers & $\left[1, 2, 3, 4\right]$ \\
        & VAE-Learning rate & LogUniform[0.00001, 0.003] \\
        & VAE-Epochs & $4000$ \\
        & Diffusion- MLP denoising dim & $\left[512, 1024, 2048\right]$ \\
        & Diffusion-batch size & $\left[512, 1024, 2048, 4096\right]$ \\
        & Diffusion-learning rate & LogUniform[0.00001, 0.003] \\
        & Diffusion-Epochs & $10000$ \\
        \hline
        
        \multirow{5}{*}{TabDDPM} & timesteps & $\left[100, 200, 300, 400, 600, 800, 1000\right]$ \\
        & latent space size & $\left[64, 128, 256, 512, 1024, 2048, 4096\right]$ \\
        & mlp depth & $\left[2, 4, 6, 8\right]$ \\
        & batch size & $\left[64, 128, 256, 512, 1024\right]$ \\
        & learning rate & LogUniform[0.00001, 0.003] \\
        & epochs & $500$ \\
        \hline
        
        \multirow{7}{*}{\modelname{}} & timesteps & $\left[100, 200, 300, 400, 600, 800, 1000\right]$ \\
        & latent space size & $\left[64, 128, 256, 512\right]$ \\
        & transformer layer num & $\left[2, 3, 4\right]$ \\
        & transformer heads & $\left[2, 4, 8\right]$ \\
        & transformer feedforward size & $\left[256, 512\right]$ \\
        & batch size & $\left[64, 128, 256, 512, 1024\right]$ \\
        & learning rate & LogUniform[0.00001, 0.003] \\
        & epochs & $500$ \\
        \bottomrule
        
    \end{tabular}}
\end{table}

\section{Addition Experimental Results} \label{sec:addtional_results}

In this Appendix, we compare the training and sampling times, as well as the number of trainable parameters, of \textit{\modelname{}} against other tabular generative models, using the \textit{Churn}, \textit{California Housing} and \textit{Gas Concentrations} datasets as examples. The training and sampling times and the number of trainable parameters are highly influenced by both the model hyperparameters and the dataset characteristics (size and number of features).

Regarding model hyperparameters, each model has been optimized for these datasets as outlined in Section \ref{sec:syn_data_gen}. This process involves using Bayesian optimization and Optuna to identify the best hyperparameters for each model, enhancing the machine learning efficiency described in Section \ref{sec:ML_eff}. The specific hyperparameter search space for each model is provided in Table \ref{tab:model_search_space} in Appendix \ref{sec:hyperparameter}, and Table \ref{tab:w_t} presents the training and sampling times, along with the number of trainable parameters, for the optimized models on the \textit{Churn}, \textit{California Housing} and \textit{Gas Concentrations} datasets.

In terms of dataset characteristics, \textit{Churn}, \textit{California Housing} and \textit{Gas Concentrations} differ in size ($10000$ vs. $20640$ vs. $13910$ samples) and feature count ($10$ vs. $8$ vs. $129$), impacting absolute values such as training and sampling times, as well as the number of trainable parameters. Despite these differences, all the datasets yield consistent results.

\begin{table}[H]
\caption{\textit{Taining/Sampling time} and \textit{number of trainable parameters} comparison for \textit{Churn}, \textit{California Housing} and \textit{Gas Concentrations} datasets.}
\label{tab:w_t}
\resizebox{0.95\textwidth}{!}{
\begin{tabular}{l|ccc|ccc|ccc}
    \toprule
     & \multicolumn{3}{c|}{Churn} & \multicolumn{3}{c|}{California Housing} & \multicolumn{3}{c}{Gas Concentrations} \\ \cmidrule(lr){2-4} \cmidrule(lr){5-7}\cmidrule(lr){8-10}
     & Training & Sampling & Parameters   & Training & Sampling & Parameters & Training & Sampling & Parameters   \\ 
    \midrule
    TVAE             & 5m 3s     & 0.86s  &  190122 & 8m 10s      & 1.96s   &  180132 & 14m 47s & 4.35s & 325108 \\
    CTGAN            & 14m 25s   & 1.68s  &  280121 & 18m 36s     & 3.01s   &  265132 & 29m 58s & 5.21s & 456587 \\
    CoDi             & 2h 26m 5s & 6.64s  &  615212 & 2h 54m 10s  & 10.54s  &  567987 & 3h 58m 15s & 16.38s & 953578 \\
    Tabsyn           & 31m 34s   & 6.02s  &  415122 & 38m 42s     & 9.16s   &  409876 & 1h 5m 31s & 14.43s & 762463 \\
    TabDDPM          & 14m 36s   & 12.75s &  221363 & 18m 12s     & 18.21s  &  209815 & 29m 36s & 28.48s & 420589 \\
    \modelname{}~I   & 14m 42s   & 15.12s &  332049 & 18m 10s     & 21.10s  &  298844 & 31m 48s & 30.45s & 597688 \\
    \modelname{}~II  & 14m 53s   & 15.30s &  332049 & 18m 42s     & 22.95s  &  298844 & 32m 24s & 35.12s & 597688 \\
    \bottomrule
\end{tabular}}
\end{table}

From the results in Table \ref{tab:ml_res}, the baseline model that most closely matches the performance of \textit{\modelname{}~I} and \textit{\modelname{}~II} in terms of machine learning efficiency is \textit{Tabsyn}. \textit{Tabsyn} is a latent diffusion model where a VAE first projects tabular data into a dense, homogeneous continuous space, followed by a diffusion process. Compared to our model, \textit{Tabsyn} has a greater number of trainable parameters and longer training times. However, in terms of sampling, \textit{Tabsyn} outperforms our model. This advantage largely arises from the findings of \cite{Karras2024Elucidating}, which suggest that the sampling process can be accelerated by reducing the number of backward diffusion steps through an effective choice of schedule and scale functions (see \cite{Karras2024Elucidating} for details). The schedule function determines the desired noise level at each diffusion time step, while the scale function dictates how data scales over time. Their optimal selection aligns with that proposed by DDIM \cite{song2021denoising}. As highlighted by \cite{Karras2024Elucidating}, this choice is independent of how the denoising model (e.g., the neural network) has been trained. Therefore, we plan to investigate the integration of our masking training mechanism with the sampling approach proposed by \cite{Karras2024Elucidating} in future work to improve the model’s sampling speed.

A final comment addresses the impact of dynamic conditioning, specifically the differences between \textit{\modelname{}~I} and \textit{\modelname{}~II}. Table \ref{tab:w_t} demonstrates that the conditioning mechanism has a limited effect on sampling and training time, while the number of trainable parameters remains unchanged with or without conditioning.

This outcome arises because the conditioning mask only influences the input size of the encoder-decoder transformer, specifically affecting the sequence length of the transformer encoder input (as illustrated in Figure \ref{fig:transformer} and discussed in Section \ref{sec:sampling}). Notably, the number of trainable parameters in a transformer encoder does not depend on input length; therefore, \textit{\modelname{}~I} and \textit{\modelname{}~II} both retain the same parameter count. Although input sequence length can impact computational time, this effect is minimal in our case. Even with the largest dataset, consisting of 129 features, the encoder input remains a relatively short sequence of 129 tokens. Consequently, the impact on both sampling and training time is minimal.

\end{document}